%% file: 0_main.tex
\documentclass[sigconf, 10pt]{acmart}

\usepackage{caption}
\usepackage{subcaption}
\usepackage{color}
\usepackage{makecell}
\usepackage{tabularx}
\usepackage{multirow}
\usepackage{multicol}
\usepackage{float}
\usepackage{enumitem}
\usepackage{graphicx}

\begin{document}

\acmYear{2024}\copyrightyear{2024}
\setcopyright{acmlicensed}
\acmConference[ACM MobiCom '24]{International Conference On Mobile Computing And Networking}{September 30--October 4, 2024}{Washington D.C., DC, USA}
\acmBooktitle{International Conference On Mobile Computing And Networking (ACM MobiCom '24), September 30--October 4, 2024, Washington D.C., DC, USA}
\acmDOI{10.1145/3636534.3649370}
\acmISBN{979-8-4007-0489-5/24/09}

\title{ADMarker: A Multi-Modal Federated Learning System for Monitoring Digital Biomarkers of Alzheimer's Disease
}

\author{Xiaomin Ouyang$^{1}$, Xian Shuai$^{1}$, Yang Li$^{1}$, Li Pan$^{1}$, Xifan Zhang$^{1}$, Heming Fu$^{1}$, Sitong Cheng$^{1}$, Xinyan Wang$^{1}$, Shihua Cao$^{1}$, Jiang Xin$^{1}$, Hazel Mok$^{1}$, Zhenyu Yan$^{1}$, Doris Sau Fung Yu$^{2}$, Timothy Kwok$^{1}$, Guoliang Xing$^{1,*}$}

\affiliation{  
\institution{$^{1}$The Chinese University of Hong Kong, $^{2}$The University of Hong Kong
\country{}
}
}

\thanks{*Corresponding author.}

\renewcommand{\shortauthors}{X. Ouyang, et. al}
\renewcommand{\shorttitle}{ADMarker: A Multi-FL System for Monitoring Digital Biomarkers of AD}

\begin{abstract}
    Alzheimer's Disease (AD) and related dementia are a growing global health challenge due to the aging population. In this paper, we present ADMarker, the first end-to-end system that integrates multi-modal sensors and new federated learning algorithms for detecting multidimensional AD digital biomarkers in natural living environments. ADMarker features a novel three-stage multi-modal federated learning architecture that can accurately detect digital biomarkers in a privacy-preserving manner. Our approach collectively addresses several major real-world challenges, such as limited data labels, data heterogeneity, and limited computing resources. We built a compact multi-modality hardware system and deployed it in a four-week clinical trial involving 91 elderly participants. The results indicate that ADMarker can accurately detect a comprehensive set of digital biomarkers with up to 93.8\% accuracy and identify early AD with an average of 88.9\% accuracy. ADMarker offers a new platform that can allow AD clinicians to characterize and track the complex correlation between multidimensional interpretable digital biomarkers, demographic factors of patients, and AD diagnosis in a longitudinal manner. 
\end{abstract}



\maketitle

\input{1_introduction.tex}

\input{2_related_work.tex}

\input{3_overview.tex}

\input{4_FL_design}

\input{5_medical}
\input{6_implementation}

\input{7_deployment.tex}
\input{8_experiments.tex}
\input{9_discussion}
\input{10_conclusion}

\begin{acks}
This work is supported by the Research Grants Council (RGC) of Hong Kong, China, under grants CRF C4034-21G and GRF 14212323, and the Alzheimer's Drug Discovery Foundation, under Grant RDADB-201906-2019049.
\end{acks}

\bibliographystyle{ACM-Reference-Format}
\bibliography{reference}

\end{document}

%% file: 1_introduction.tex
\section{Introduction}

Alzheimer's Disease (AD) is a progressive neurodegenerative disease that can cause significantly declining cognitive and functional abilities. AD and related dementia is a growing health challenge worldwide because of population aging \cite{livingston2017dementia, livingston2020dementia}. In 2010, about 35.6 million people lived with dementia, which is expected to double every 20 years \cite{fiest2016prevalence}. 

A major barrier to the treatment of AD is that many patients are either not diagnosed or diagnosed at the late stages of the disease. Studies suggest that 75\% of worldwide persons with dementia are undiagnosed \cite{ad-report-diagnosis}. This is largely due to the fact that, the standard clinical procedure for AD diagnosis, based on Magnetic Resonance Imaging (MRI), Positron Emission Tomography (PET) brain scan, or blood biomarkers like amyloid $\beta_{1-42}$, is only available in clinical settings. Although various screening tests can identify cognitive impairments, they are usually intrusive and cannot be conducted routinely or in a real-time manner. Therefore, early identification of people at risk of developing AD and timely intervention to slow the onset and progression of AD are crucial.

 A recent major advance in early AD diagnosis and intervention is to leverage AI and sensor devices to capture physiological, behavioral, and lifestyle symptoms of AD (e.g., activities of daily living and social interactions) in natural home environments, referred to as \emph{digital biomarkers} \cite{digital-biomarker, ADDF-xing, kourtis2019digital}. The difficulty in performing activities of daily living (ADLs) is a hallmark feature of AD, because ADLs, such as watching TV, cleaning living areas, and taking medicine, involve tasks that require independence, organization, judgment, and sequencing abilities \cite{tekin2001activities}. 
Moreover, social behaviors, such as family meals and phone calls, are shown to be strongly correlated with the risk of early AD \cite{doris2016measuring, arai2021influence}.

There are several major challenges that have not been addressed in previous work on AD digital biomarkers. First, existing work is focused on a particular type of digital biomarker, such as motor function \cite{li2018tatc}, speech features \cite{nasreen2021detecting}, or driving habits \cite{bayat2021gps}, which lacks generalizability to subjects with various demographic and medical characteristics. Second, most of the studies are based on the black-box approach, where the sensor data/feature is directly used to identify AD, which not only incurs extremely high computing overhead but also results in digital biomarkers that are difficult to interpret by medical professionals. {Third, existing solutions are based on a centralized learning approach that needs to upload raw sensor data, imposing significant privacy concerns.}


\begin{table*}
    \centering
 \setlength{\abovecaptionskip}{0.cm}
    \setlength{\belowcaptionskip}{0.cm}
    \resizebox{\linewidth}{!}{
     \begin{tabular}{ccccccc}
      \toprule
       & Approach & Accuracy & \makecell[c]{\# of biomarkers} & \makecell[c]{Sensors} & \makecell[c]{\# of subjects} &  \makecell[c]{Duration}\\
      \midrule
     Tatc \cite{li2018tatc} & Black-box, Centralized & 42.3\% for MCI & 1 (sensor data) & Actigraphy & 729 (185 AD, 103 MCI, 441 NC)  & 7 days \\
     ADReSS \cite{luz2020alzheimer} & Black-box, Centralized & 60.8\% for AD & 6 acoustic features & Audio & 156 (78 AD, 78 non-AD)  & 10mins (in lab)\\
     Bayat et al. \cite{bayat2021gps} & Black-box, Centralized & 82\% fro AD & 14 GPS driving indicators & In-vehicle GPS & 139 (64 AD, 75 non-AD) & One year\\
     Alberdi et al. \cite{alberdi2018smart} & Interpretable, Centralized & No diagnosis results & Features of 5 events & PIR motion sensor & 29 (6 AD, 10 MCI, 12 NC) & Two years\\
     ADMarker (Ours) & \textbf{Interpretable, Distributed} & \textbf{88.9\%} for MCI & Features of \textbf{22} activities &  Audio, Depth, Radar & 91 (31 AD, 30 MCI, 30 NC) & Four weeks\\
      \bottomrule
    \end{tabular}}
    \caption{Comparison of different AD digital biomarker studies. ADMarker is the first solution that detects a comprehensive set of multidimensional digital biomarkers for AD diagnosis in a privacy-preserving manner.} 
    \label{table:comparison_digital_biomarker}
    \vspace{-1em}
\end{table*}


In this paper, we present ADMarker, the first end-to-end system that integrates multi-modal sensors and new federated learning (FL) algorithms for detecting multidimensional, more than 20 AD digital biomarkers in a privacy-preserving manner. The system features a novel three-stage FL architecture, where the nodes deployed in subjects' homes leverage the pre-trained model to reduce the compute overhead of online training, and improve the model performance on their own data through multi-modal unsupervised and weakly supervised FL algorithms. This approach collectively addresses several real-world challenges, including limited labeled data, data heterogeneity, and limited computing resources. 
We implemented ADMarker on a compact multi-modality sensor hardware system with three privacy-preserving sensors (i.e., a depth camera, a mmWave radar, and a microphone) to detect a comprehensive set of digital biomarkers in home environments.
The design of ADMarker also addresses several practical challenges to ensure the durability and efficiency of the hardware and software system, including long-term multi-modal data recording, improving training and inference efficiency, and private communication networks. 

We have deployed ADMarker in a four-week clinical trial that involves a total of 91 elderly subjects, including 31 with AD and 30 with mild cognitive impairment (MCI). {
After excluding 22 subjects with very limited valid data, we evaluate the performance of ADMarker on the data of 69 remaining subjects. The results show that,}
ADMarker can detect more than 20 daily activities in natural home environments with up to 93.8\% detection accuracy, using only a very small amount of labeled data. Leveraging the detected digital biomarkers, we can achieve 88.9\% accuracy in early AD diagnosis. ADMarker offers a new clinical tool that allows medical researchers and professionals to monitor the progression of AD manifestations and study the complex correlation between multidimensional interpretable digital biomarkers, demographic factors of patients, and AD diagnosis in a longitudinal manner.

{
In summary, our key contributions include:
\begin{itemize}[noitemsep,leftmargin = *]
    \item We identify a comprehensive set of digital biomarkers that are strongly associated with different AD stages yet can be detected during daily life, and develop a compact multi-modal hardware system that can be rapidly deployed in home environments to detect these biomarkers.
    \item We propose a novel three-stage multi-modal federated learning approach that incorporates novel unsupervised and weakly supervised multi-modal FL designs to collectively address several major real-world challenges in biomarker detection, including limited data labels, data heterogeneity, and limited computing resources. 
    \item We deployed our systems in a four-week clinical trial involving 91 elderly participants. The results show that ADMarker can accurately detect a comprehensive set of digital biomarkers and identify early AD with high accuracy.
\end{itemize}
}


%% file: 2_related_work.tex
\section{Related Work}
\label{sec:related_work}

\textbf{Mobile Health Systems.} 
Numerous mobile health systems have been proposed in recent years \cite{silva2015mobile,istepanian2007m,isleep,stresssense}. 
Recently, several mobile systems are developed for the assessment or rehabilitation of neurodegenerative diseases. 
For example, NeuralGait \cite{NeuralGait} captures the gait segments relationship for brain health assessment. PDlens \cite{pdlens} uses smartphone data for drug-effectiveness detection of Parkinson's disease. However, these systems can only monitor very limited symptoms of the disease, e.g., the gait or sound of subjects. ADMarker is the first system that can detect multi-dimensional AD digital biomarkers in a real-time and privacy-preserving manner.

\textbf{Federated Learning for Human Activity Recognition.}
Most work in Human Activity Recognition (HAR) is based on centralized learning that needs to train the algorithms at a central server \cite{stisen2015smart, jiang2018towards, xu2021limu}, which imposes significant privacy concerns due to the need to share raw user data.
Federated learning (FL) has been recently applied to HAR to improve the model accuracy without sharing the raw data \cite{ouyang2021clusterfl,ouyang2022clusterfl,tu2021feddl}. 
{However, most of the existing FL approaches are focused on training \emph{unimodal} models with a single type of sensor data modality, such as image \cite{ouyang2022clusterfl} or inertial sensory data \cite{tu2021feddl}. 
Although several multi-modal FL schemes \cite{zhao2022multimodal, ouyang2023harmony} allow model training over distributed multi-modal data on the nodes, they are focused on classifying a small number of activities in controlled environments and require labeled data from users. 
}
ADMarker is the first multi-modal FL system that incorporates novel unsupervised and weakly supervised multi-modal FL designs to detect a comprehensive set of human activities in natural living environments. 




\textbf{Digital Biomarkers for Alzheimer's Disease.} 
Recently, understanding AD using mobile sensor-based digital biomarkers has attracted tremendous attention \cite{li2018tatc, chen2019developing, luz2020alzheimer, bayat2021gps}.  
Table \ref{table:comparison_digital_biomarker} compares several representative AD digital biomarker studies. In summary, these studies have several key major issues. {First, existing works only focus on a particular type of digital biomarkers \cite{li2018tatc,luz2020alzheimer} or detect a very small number of simple activities like sitting and sleeping \cite{alberdi2018smart},} which are only applicable to partial population demographics or environments. 
Second, most of the studies are based on the black-box approach that directly infers the diagnosis from raw sensor data/features. Such an approach incurs extremely high computing overhead, as the amount of collected data over time can be huge (e.g., about 90T in our four-week clinical deployment). In addition, even if the diagnosis with raw sensor data is accurate, the results are difficult to interpret and adopt in current practices of AD diagnosis and treatment that are largely based on observable cognitive and behavioral symptoms \cite{digital-biomarker, kourtis2019digital}. {Third, most existing solutions adopt a centralized learning approach. ADMarker is the first system that can detect behavior biomarkers in a distributed learning manner to preserve users’ data privacy.}

%% file: 3_overview.tex
\begin{figure*}
    \setlength{\abovecaptionskip}{-0.1cm}
    \setlength{\belowcaptionskip}{0.cm}
    \centering
     \includegraphics[width = \linewidth]{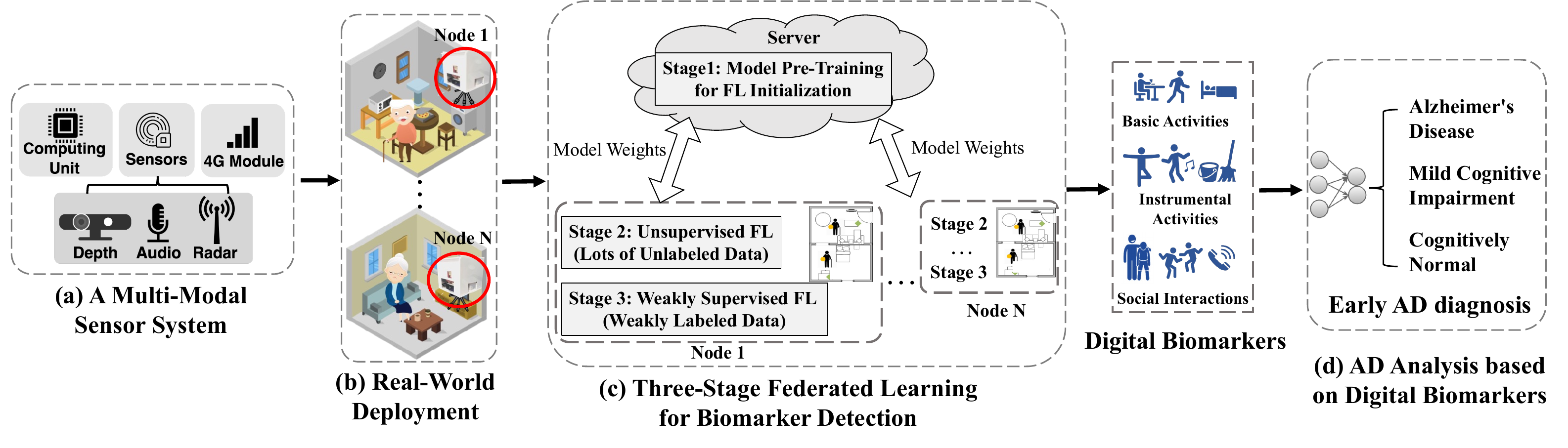}
  \caption{Overview of ADMarker. ADMarker consists of three major components, i.e., a multi-modal sensor system, federated learning for biomarker detection, and AD analysis based on detected digital biomarkers.}
  \label{fig:system-overview}
  \vspace{-0.5em}
\end{figure*}

\section{System Overview}
\label{sec:overview}



\subsection{Motivation}

\textbf{Interpretable two-step AD monitoring.} 
Instead of predicting AD with the raw multi-modal sensor data (i.e., black-box approaches in Table \ref{table:comparison_digital_biomarker}), ADMarker disentangles the disease monitoring into two steps, i.e., digital biomarkers detection and early disease analysis with the detected biomarkers. As a result, the behavior biomarkers detected by ADMarker are more interpretable for AD diagnosis and can be adopted to design personalized intervention plans. {For example, we select a large set of biomarkers that are not only strongly associated with different stages of AD (see Table \ref{table:task_index}), but also consistent with the current practice of AD intervention. 
Moreover, we could identify critical digital biomarkers through correlation analysis between the biomarkers and diagnosis results.
}




\textbf{Detecting multi-dimensional biomarkers.} 
Compared with existing approaches focused on a particular type of digital biomarker \cite{li2018tatc, luz2020alzheimer, bayat2021gps}, ADMarker aims to detect a comprehensive set of multi-dimensional AD symptoms, such as various activities of daily living and social interactions. This allows to investigate how the physical and social interactional etiopathogenesis shape AD's manifestation. Furthermore, the integration of multi-dimensional digital biomarkers enables the precise detection of early AD, encompassing its diverse manifestations, which would be challenging when relying solely on a single type of biomarker \cite{livingston2017dementia,livingston2020dementia}.

\textbf{Federated learning for privacy-preserving biomarker detection.} To continuously and accurately detect the biomarkers, ADMarker needs to accumulate data for a certain period and use it to train ML models for daily activity recognition. 
In order to preserve users' data privacy, ADMarker {avoids uploading raw sensor data through federated learning \cite{konevcny2016federated1, ouyang2021clusterfl} and communicates with the server via secured networks.} 



\subsection{Challenges}
\label{sec:challenges}
{The design of ADMarker (including the hardware, software, algorithm, and clinical protocol) requires extensive  expertise and efforts in both medical and engineering domains. Through our clinical deployment, we identify many real-world challenges that haven’t been adequately addressed and collectively tackle them via both system and algorithm designs.} 
Specifically,
ADMarker aims to address three major challenges.

The first challenge is to \emph{utilize a large set of digital biomarkers for early AD diagnosis and intervention}. It is essential to select a comprehensive set of activities that are highly associated with different AD stages yet can be detected accurately during daily life. This requires extensive domain expertise in both medical and engineering areas. 

The second challenge is to develop \emph{a hardware system that can be rapidly deployed in real-world home environments} for longitudinal daily activity monitoring. For instance, in order to capture multi-dimensional digital biomarkers in a privacy-preserving manner, the hardware system should incorporate carefully selected sensors of different modalities. 


The third challenge is to design \emph{an effective multi-modal federated learning system that can accurately detect digital biomarkers} under real-world data and system dynamics. First, there usually exists no labels or only a very limited amount of labeled data because most sensor data is not intuitive for humans to label \cite{shuai2021millieye}.
Second, due to the significant diversity in behavior patterns and home environments, the data distribution of different subjects is usually non-i.i.d. and highly imbalanced. 
Finally, there may be significant training latency in real-world FL systems due to the limited computing resources and dynamic bandwidth of nodes.



\subsection{System Architecture}

Our key idea is to leverage multi-modal sensor devices and federated learning algorithms to detect multi-dimensional AD digital biomarkers in natural home environments. Figure~\ref{fig:system-overview} shows the overview of ADMarker.

We developed a compact multi-modality hardware system that can function for up to months in home environments to detect digital biomarkers of AD. It incorporates three privacy-preserving sensors (a depth camera, an mmWave radar, and a microphone), an NVIDIA single-board edge computer, and a 4G cellular interface that can communicate with the server. We address several practical challenges to make the hardware system durable, power-efficient, and privacy-preserving. On top of the hardware system, we design a new multi-modal federated learning (FL) system for biomarker detection while preserving users' data privacy. The system features a novel three-stage architecture. At the first stage, we train a multi-modal model on the server using the labeled data from public datasets or the previous participants. Then, during the deployment, the nodes in the subjects' homes will load the centrally pre-trained model, and improve the model performance on their own data through a two-stage FL process, i.e., multi-modal unsupervised and weakly supervised FL, respectively. 
In weakly supervised FL, the nodes will only leverage \emph{weak labels} generated from sparse activity logs of the participants, without resorting to labor-intensive manual annotation of raw sensor data, for local model training.
Our approach collectively addresses several major real-world challenges, including limited labeled data, data heterogeneity, and limited computing resources. Finally, the digital biomarkers are input into a neural network for AD diagnosis of different patients.

The design of ADMarker is extensively evaluated in a clinical deployment (see Section \ref{sec:deployment}). A total of 91 elderly subjects (43 females and 48 males, 61-93 years old) were recruited for the study, including 31 with AD, 30 with mild cognitive impairment, and 30 cognitively normal subjects. 

\subsection{Biomarker Selection} 
We select a total of 22 activities of interest that are shown to be highly related to AD from medical literature \cite{berkman2000social, marique2005cerebral, marshall2006neuropathologic, altieri2021functional}. ADMarker will detect these activities and use the duration and frequency of the activities as potential digital biomarkers for diagnostic analysis (see Section~\ref{sec:medical_analysis}).

\begin{table}
\centering
\renewcommand{\arraystretch}{1} 
\scriptsize


\begin{tabular}{>{\centering\hspace{0pt}}m{0.03\linewidth}>{\centering\hspace{0pt}}m{0.29\linewidth}>{\centering\hspace{0pt}}m{0.22\linewidth}>{\centering\hspace{0pt}}m{0.08\linewidth}>{\centering\arraybackslash\hspace{0pt}}m{0.21\linewidth}}

\toprule
{Class\par{}Index} & Activities & Captured\par{}Sensors & Type & {Decline at\par{}which stage}\\ 
\midrule
1 & Out of Home & Depth, Radar, Mic &  &  Mild, Moderate\\ 
\hline
2 & Other activities & Depth, Radar, Mic &  &  \\ 
\hline
3 & Dressing & Depth, Radar & BADL & Moderate, Severe \\ 
\hline
4 & Take/Put something & Depth, Radar & BADL & Severe \\ 
\hline
5 & Cleaning living area & Depth, Radar, Mic & IADL & Mild \\ 
\hline
6 & Grooming & Depth, Radar & IADL & Moderate, Severe \\ 
\hline
7 & Wiping hands & Depth, Radar & BADL & Severe \\ 
\hline
8 & Drinking & Depth, Radar & BADL & Moderate, Severe \\ 
\hline
9 & Eating & Depth, Radar, Mic & BADL & Moderate, Severe \\ 
\hline
10 & Smoking & Depth, Radar & IADL & Mild \\ 
\hline
11 & Sneezing/Coughing & Depth, Radar, Mic & BADL & Mild\\ 
\hline
12 & Writing & Depth, Radar & IADL & Mild \\ 
\hline
13 & Watching TV & Depth, Radar, Mic & BADL & Mild \\ 
\hline
14 & Phone call/Using phone & Depth, Radar, Mic & SI & Mild \\ 
\hline
15 & Exercising & Depth, Radar & IADL & Moderate, Severe \\ 
\hline
16 & Talking with others & Depth, Radar, Mic & SI & Severe \\ 
\hline
17 & Stretching & Depth, Radar & BADL & Mild, Moderate \\ 
\hline
18 & Walking & Depth, Radar & BADL & Moderate, Severe \\ 
\hline
19 & Sitting & Depth, Radar & BADL & Moderate, Severe \\ 
\hline
20 & Standing & Depth, Radar & BADL & Severe \\ 
\hline
21 & Lying & Depth, Radar & BADL & Moderate, Severe \\ 
\hline
22 & Moving in/out of chair & Depth, Radar & IADL & Severe \\
\bottomrule
\end{tabular}
\caption{ Selected AD digital biomarkers. BADL: Basic Activities of Daily Living; IADL: Instrumental Activities of Daily Living; SI: Social Interaction.}
\label{table:task_index}
\vspace{-3.5em}
\end{table}

Table \ref{table:task_index} shows the activities in the biomarker set, which can be categorized into three types: \emph{basic activities of daily living (BADLs)}, \emph{instrumental activities of daily living (IADLs)}, and \emph{social interactions (SI)}. 
BADLs include basic self-care tasks, such as eating, drinking, and walking, while IADLs encompass complex tasks that allow for independent living, like cleaning the living area and grooming \cite{altieri2021functional,reed2016identifying}. AD patients who suffer from decreased ability in ADLs and social interactions are more likely to exhibit amyloid plaques and neurofibrillary tangles in several brain regions \cite{marshall2006neuropathologic, marique2005cerebral}. 

Moreover, the last column in Table \ref{table:task_index} shows the stages of the disease where these activities become difficult for the patients. At the mild stage of AD, activities like watching TV \cite{gustafsdottir2015watching} and using phone \cite{yamada2021combining} become challenging due to early signs of cognitive decline. When the disease progresses to moderate and severe stages, the patients will suffer comprehensive functional deterioration, and are difficult to perform basic activities like standing \cite{montero2012gait}, exercising\cite{miu2008randomised}, and talking with others \cite{mcnamara1992speech}.
Therefore, the digital biomarkers selected above are not only interpretable with respect to AD manifestations, but also can be readily applied in intervention plans. For example, by monitoring the subjects’ daily activities during the progression of AD, the duration and intensity of exercise can be prescribed for personalized intervention, which leads to iterative and more effective treatment.

%% file: 4_FL_design.tex
\section{Federated Learning for Biomarker Detection}
\label{sec:fl_design}

\subsection{A Motivation Study}

\subsubsection{Understanding the real-world challenges.} We first analyze the sensor data and system log recorded by our ADMarker testbed (see Section \ref{sec:implementation}) in a four-week clinical deployment to understand the real-world challenges.



\begin{figure}
    \setlength{\abovecaptionskip}{0.cm}
    \setlength{\belowcaptionskip}{0.cm}
    \centering
     \includegraphics[width = 0.9\linewidth]{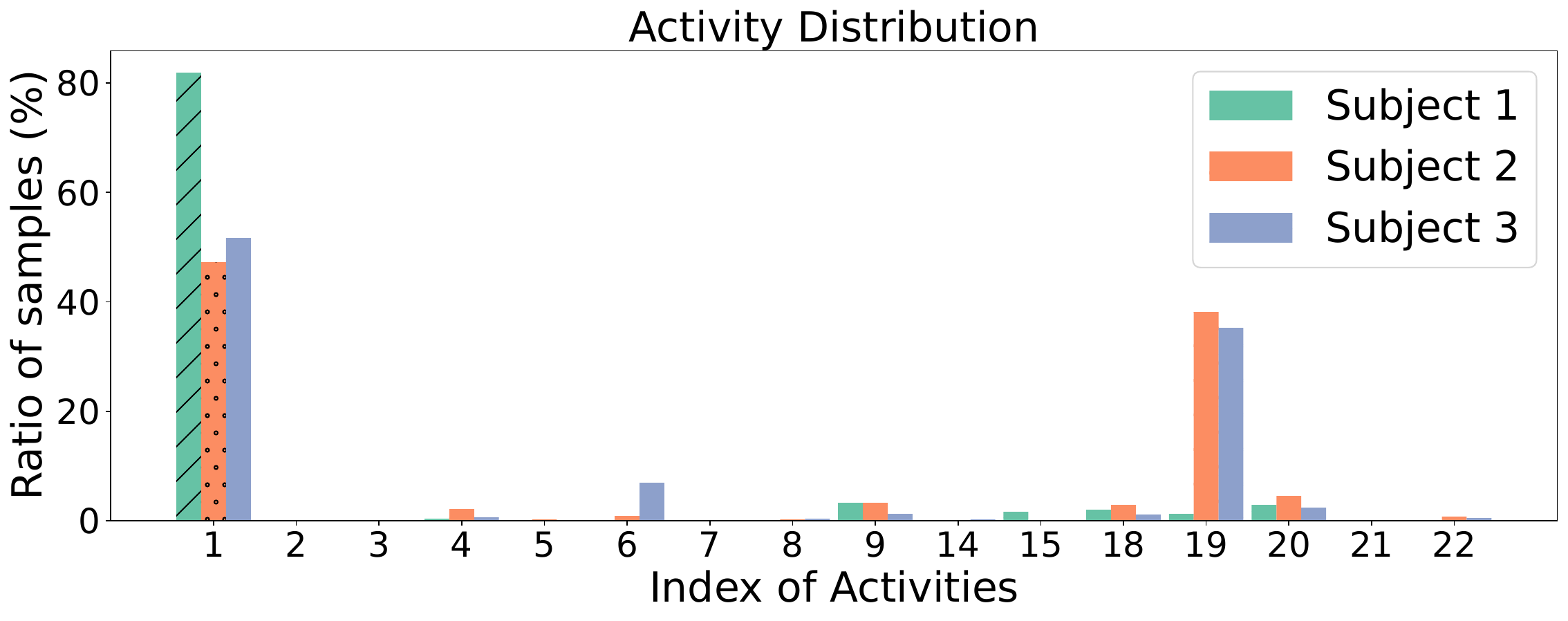}
  \caption {Examples of data distributions of participants in a real-world deployment. The data distribution is highly imbalanced among different classes and non-i.i.d across different participants.}
  \label{fig:data-distribution}
  \vspace{-1em}
\end{figure}

\textbf{Data challenges.} Figure~\ref{fig:data-distribution} shows the examples of data distributions of participants over the four weeks of the clinical deployment (see Section \ref{sec:deployment}). 
First, the amount of data from different activities is highly imbalanced. For example, in the daily living activities of Subject 3, the ratio of samples from class 1 (Out of Home, 81.78\%) is very large, while that from class 15 (Exercising, 1.54\%) is small. 
The imbalanced activity distribution will lead to a model bias on head classes, and a significant classification accuracy drop on minority tail classes \cite{liu2008exploratory,balancefl}. Second, the distributions of different subjects' data are non-i.i.d. For example, the number of occurred activities and their ratios are very different between Subject 1 (cognitively normal) and Subject 3 (AD). The non-i.i.d data distribution across nodes will reduce the accuracy performance during FL \cite{ouyang2021clusterfl, ouyang2022clusterfl}. For example, the accuracy of federated learning reduces by up to 55\% for models trained for highly skewed non-i.i.d data \cite{zhao2018federated}.
Finally, there usually exists no labels or only a very small amount of labeled data in real-world settings because the sensor data (depth images, radar point cloud, and MFCC audio features in ADMarker) is not intuitive for humans to label (as shown in Figure~\ref{fig:ADBox-data}).





\begin{figure}
    \setlength{\abovecaptionskip}{-0.1cm}
    \setlength{\belowcaptionskip}{0.cm}
    \centering
     \includegraphics[width = 0.9\linewidth]{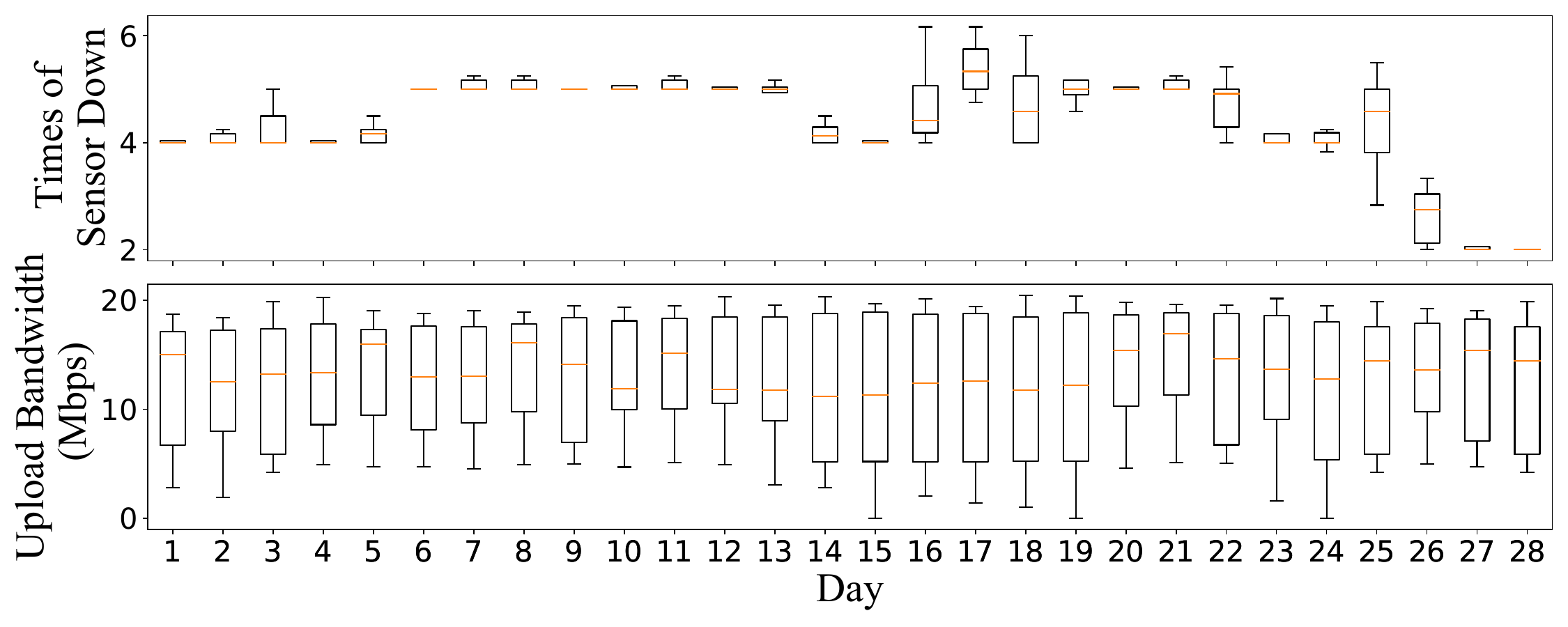}
  \caption {Times of sensor down and bandwidth of all nodes over four weeks of the real-world deployment. }
  \label{fig:system-log}
  \vspace{-1em}
\end{figure}

\label{sec:system_dynamics}
\textbf{System dynamics.} We then evaluate the long-term system dynamics using the system log recorded during the real-world deployment. The results are shown in Figure~\ref{fig:system-log}. The upper sub-figure shows the down times of sensors (per day) of the ADMarker prototype over four weeks, where each box shows the statistics of all nodes in the day. 
During the 4-week deployment, the sensors may stop data recording occasionally (about 2-6 times per day) due to the system dynamics, such as power surges or unstable sensor connections, resulting in heterogeneous sensor modalities across different nodes in FL. The lower sub-figure shows the upload bandwidth of nodes (with 4G LTE networks) over different days, which fluctuates in a significant dynamic range (e.g., 0-20 Mbps).
Such bandwidth dynamics will result in various communication delays in transmitting model weights in federated learning.

\subsubsection{Performance of different learning approaches.}
We then evaluate the performance of different traditional learning approaches to motivate the design of our federated learning architecture. The task is to classify 22 daily living activities related to Alzheimer's Disease using audio, depth, and radar data. The testing data is collected from a subject over four weeks, with a total of 500 samples. There are 200 labeled training samples and 4,000 unlabeled training samples from the target subjects. The pre-trained model is trained using labeled data (3,728 samples) from another nine subjects. 

 \begin{table}
    \centering
 \setlength{\abovecaptionskip}{0.cm}
    \setlength{\belowcaptionskip}{0.cm}
    \resizebox{\linewidth}{!}{
     \begin{tabular}{ccccc}
      \toprule
      Approach & Pre-trained & Supervised & Semi-supervised \\
      \midrule
      Training data & N/A  & Label & Unlabel+Label \\
      \midrule
      \makecell[c]{Testing Accuracy} & 18.75\% & 59.37\% & 80.62\%  \\
      \midrule
     \makecell[c]{Training Latency (h)} & 0 & 4.14 & 85.83 + 2.07 \\
      \bottomrule
    \end{tabular}}
    \caption{Model performance of three different learning approaches. ``Pre-trained'' directly applies a pe-trained model to the target subject. ``Supervised'' means training from scratch with limited labeled data. ``Semi-supervised'' trains the model with unlabeled and labeled data. The training latency is measured on the ADMarker prototype.}
    \label{table:performance_setting}
    \vspace{-1.5em}
\end{table}

Table \ref{table:performance_setting} shows the model testing accuracy and on-device training latency of different training schemes. First, directly applying the pre-trained model to the new subject results in a low testing accuracy (i.e., only 18.75\%), which shows that there is a huge domain gap among the data of different subjects. Second, when there is only limited labeled training data (the supervised approach), the model accuracy is still unsatisfactory, i.e., 59.37\%. And the model accuracy can be improved by unsupervised multi-modal learning that leverages large amounts of unlabeled data from the subject (the semi-supervised approach). However, when learning from scratch, unsupervised model training with large amounts of unlabeled data will incur significant computing overhead on the edge device, i.e., about 85 hours.

\subsection{Design Overview: A Three-Stage Federated Learning Architecture} 

\begin{figure}
    \setlength{\abovecaptionskip}{-0.cm}
    \setlength{\belowcaptionskip}{0.cm}
    \centering
     \includegraphics[width = \linewidth]{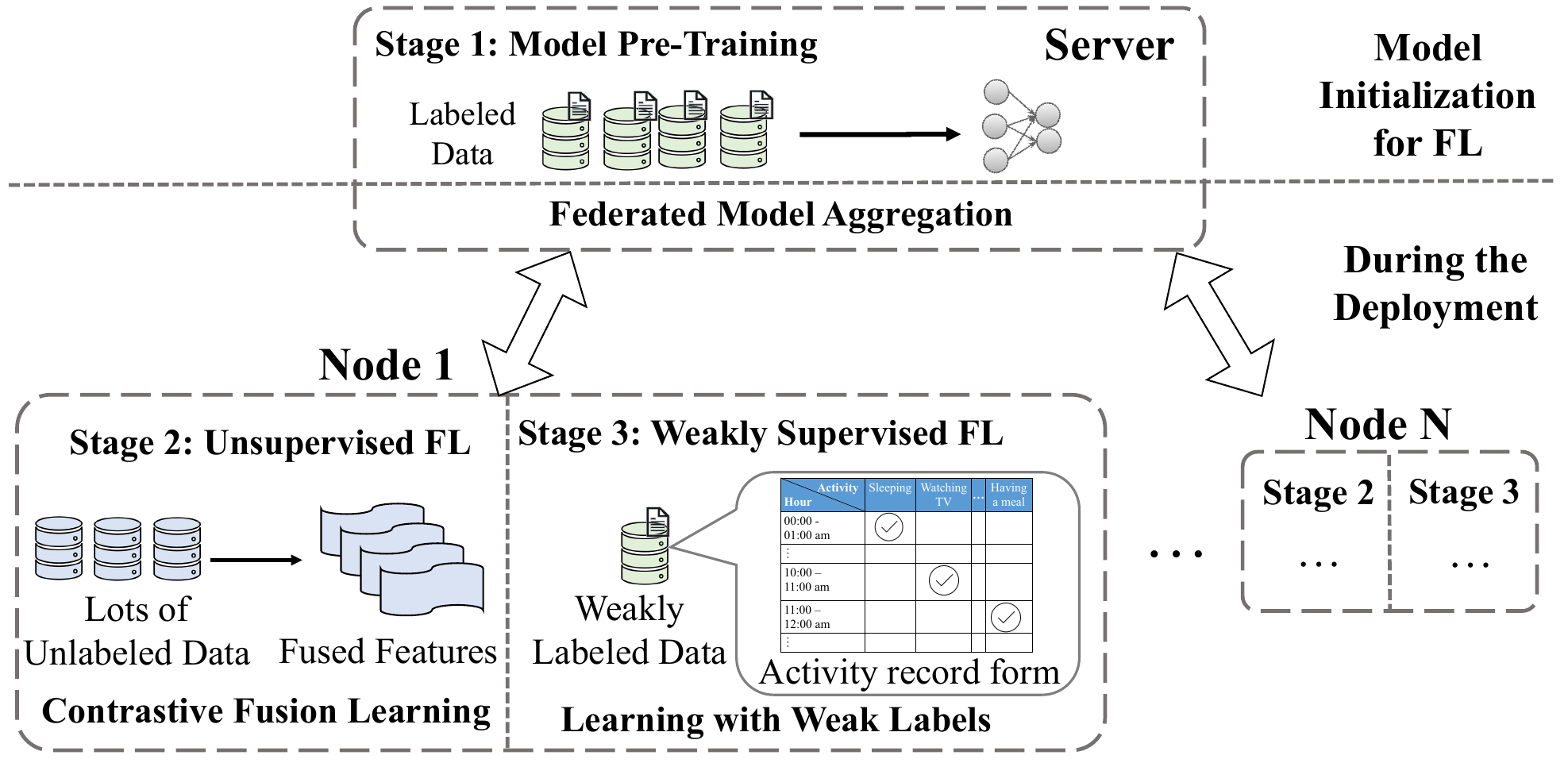}
  \caption{The three-stage federated learning architecture of ADMarker. Stage 1: Centralized model pre-training; Stage 2: Unsupervised multi-modal FL; Stage 3: Weakly supervised multi-modal FL.}
  \label{fig:three-stage-framework}
  \vspace{-2.1em}
\end{figure}

Motivated by the case study, we propose a novel three-stage federated learning architecture that integrates a pre-trained model on the cloud and leverages both unlabeled and labeled data on nodes deployed in the subjects' homes. As shown in Figure~\ref{fig:three-stage-framework}, at the first stage, we train a multi-modal model using the labeled data on the server.
Such training data can leverage the public data sets that have already been made public. For instance, the data collected from this work will be made public and utilized for model pre-training in future deployments. {We can use datasets that contain a subset of three modalities to pre-train unimodal encoders. }
Then, during the deployment, the nodes in the subjects' homes will load the centrally pre-trained model, and improve the model performance on their own data through two-stage federated learning, i.e., unsupervised multi-modal FL and weakly supervised multi-modal FL, respectively.

The reasons for the three-stage training design are as follows. First, the model trained on the cloud needs to be retrained to adapt to the local data of each subject. Second, training the multi-modal network from scratch with noisy real-world data is an extremely challenging task in FL settings. The pre-trained model can reduce the computing overhead of online FL training.
Third, there usually exists a very small amount of labeled data, as it is difficult to label multi-modal data in real-world settings. Therefore, the nodes perform unsupervised FL to leverage the large amounts of unlabeled multi-modal data, and weakly supervised FL based on limited weakly labeled data. The weak labels for supervised local training are provided by the participants (see Section \ref{sec:weak_supervise_fl}). For example, marking the time of having lunch would automatically label the multi-modal data during lunch. 
In the following, we focus on the framework of unsupervised and supervised multi-modal federated learning. 

\vspace{-0.5em}
\subsection{Unsupervised Multi-Modal FL}
In this stage, the nodes will download the pre-trained model from the server, and collaboratively train feature encoder networks of different data modalities (i.e., depth images, audio, and radar data) using the collected unlabeled sensor data during deployment. There are two main challenges during the unsupervised federated learning stage. First, the sensor modalities produce highly heterogeneous information about the same events/activities. For example, the audio features and radar data have significantly different dimensions and patterns, making it challenging to extract useful information. Second, the sensor modalities available on different nodes may vary due to the deployment constraints or runtime system dynamics. For example, some families may not be willing to have depth cameras installed in the bedroom, and the sensors may fail dynamically, e.g., due to power surges. Therefore, the design of unsupervised multi-modal FL should adapt to different modality combinations. 

To address these challenges, the nodes will run fusion-based contrastive learning that trains the feature encoders by exploring the consistent information of heterogeneous data modalities. The server will aggregate the feature encoders of nodes with heterogeneous data modalities through modality-wise federated averaging. Compared with traditional multi-modal FL approaches, the key advantage of this idea is that it is oblivious to differences of modalities on nodes.

\begin{figure}
    \setlength{\abovecaptionskip}{0.cm}
    \setlength{\belowcaptionskip}{0.cm}
    \centering
     \includegraphics[width = 0.9\linewidth]{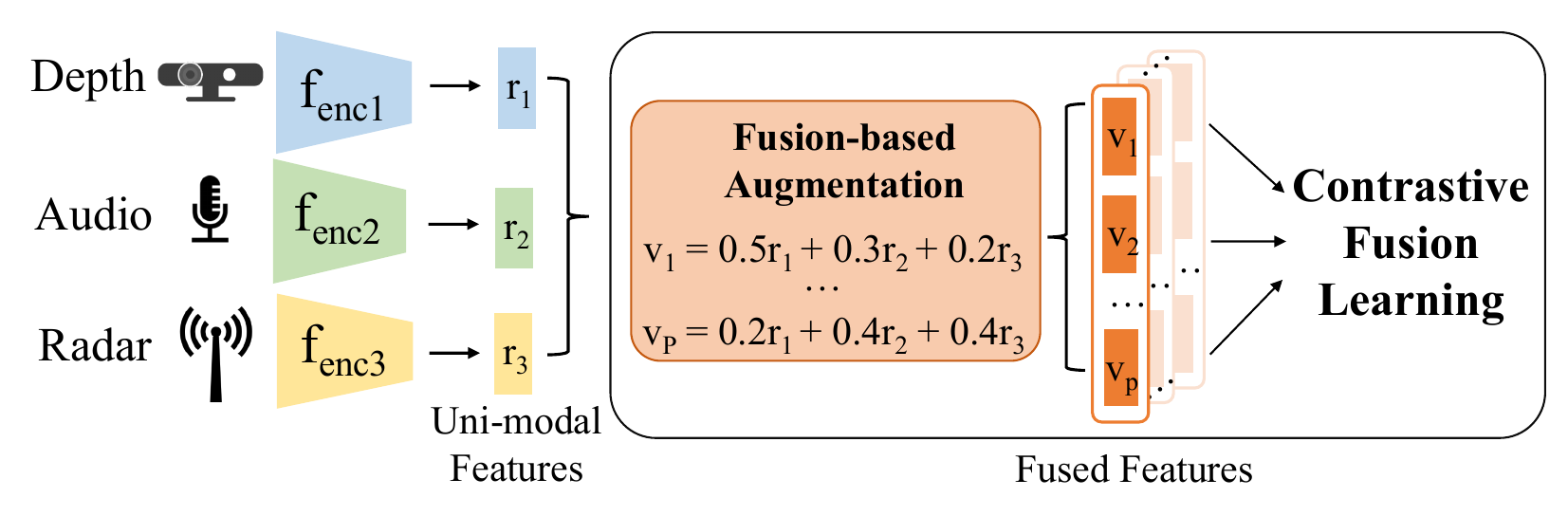}
  \caption{Contrastive fusion learning on nodes with unlabeled multi-modal data. Through contrastive learning on augmented fused features, the feature encoders are trained to capture consistent information.}
  \label{fig:stage2-contrastive}
  \vspace{-1.em}
\end{figure}

\textbf{Contrastive fusion learning on nodes.} During local training, a fusion-based feature augmentation module \cite{ouyang2022cosmo} will extensively augment the uni-modal features (extracted by encoders of different modalities) to a group of fused features via weighted sum or concatenation. 
As shown in Figure \ref{fig:stage2-contrastive}, each augmented feature represents a different fusion combination of the sensor features and contains some subset of information in the original data sample.
Let $s \in \emph{S} \equiv \{1, 2, ..., P \times N\}$ be the index of an arbitrary augmented feature, and let $p \in P(s) $ be the index of the other augmented features originating from the same source sample. 
The contrastive fusion loss can be defined as:
\begin{equation}
    \mathcal{L}_{conf} = \sum_{s \in \emph{S}} \frac{-1}{|P(s)|} \sum_{p \in P(s)}log
    \frac{exp(\mathbf{v}_s \cdot \mathbf{v}_p / \tau ) } { \sum_{a \in \emph{S} \backslash \{ s \} } exp(\mathbf{v}_s \cdot \mathbf{v}_a / \tau) }.\label{eq:confusion_loss}
\end{equation}
Here $\mathbf{v}_s$ is the feature output of the fusion-based augmentation module, and the symbol $\cdot$ denotes the inner product of feature vectors. $\tau \in \mathbb{R}^{+}$ represents the temperature used to adjust the impact of different samples \cite{tian2020makes}.
Therefore, minimizing the contrastive fusion loss will force the fused features from the same multi-modal data sample (positive features, $\mathbf{v}_s$ and $\mathbf{v}_p$) together, while pushing fused features from other data samples (negative features, $\mathbf{v}_s$ and $\mathbf{v}_a$) apart. In this way, the feature encoders are trained to learn consistent information across modalities by maximizing the mutual information of features from different modalities.

\textbf{Modality-wise federated average.} In multi-modal FL systems, the nodes with different data modalities will have different model architectures. For example, the nodes with all data modalities will train models with three multi-modal feature encoders, while the models of nodes with only depth and audio data will train models with two feature encoders. 
We propose a modality-wise federated average scheme to address the challenge of modality heterogeneity, where the server will collect and aggregate the encoder networks of the same modality with Fedavg \cite{mcmahan2016communication}. As a result, the nodes with different data modalities can collaborate to improve the performance in unsupervised federated learning. We also apply this model aggregation scheme on the server during the weakly supervised multi-modal FL stage.

 \subsection{Weakly Supervised Multi-Modal FL}
 \label{sec:weak_supervise_fl}
At the third stage, the nodes will perform weakly supervised multi-modal FL based on the model trained at the second stage. During the real-world deployment, the nodes can leverage weak labels provided by the participants or their caregivers for local model training. For example, marking the time of having lunch would automatically label the sensor data during the period. However, it would be challenging to associate the sparse and noisy labels with collected sensor data for supervised model training. Moreover, different subjects usually have highly heterogeneous and imbalanced data distributions, making it challenging for model aggregation in FL. For example, some activities such as ``sitting'' incur frequently while others like ``writing'' appear rarely. 



\begin{figure}
    \setlength{\abovecaptionskip}{0.cm}
    \setlength{\belowcaptionskip}{0.cm}
    \centering
     \includegraphics[width = \linewidth]{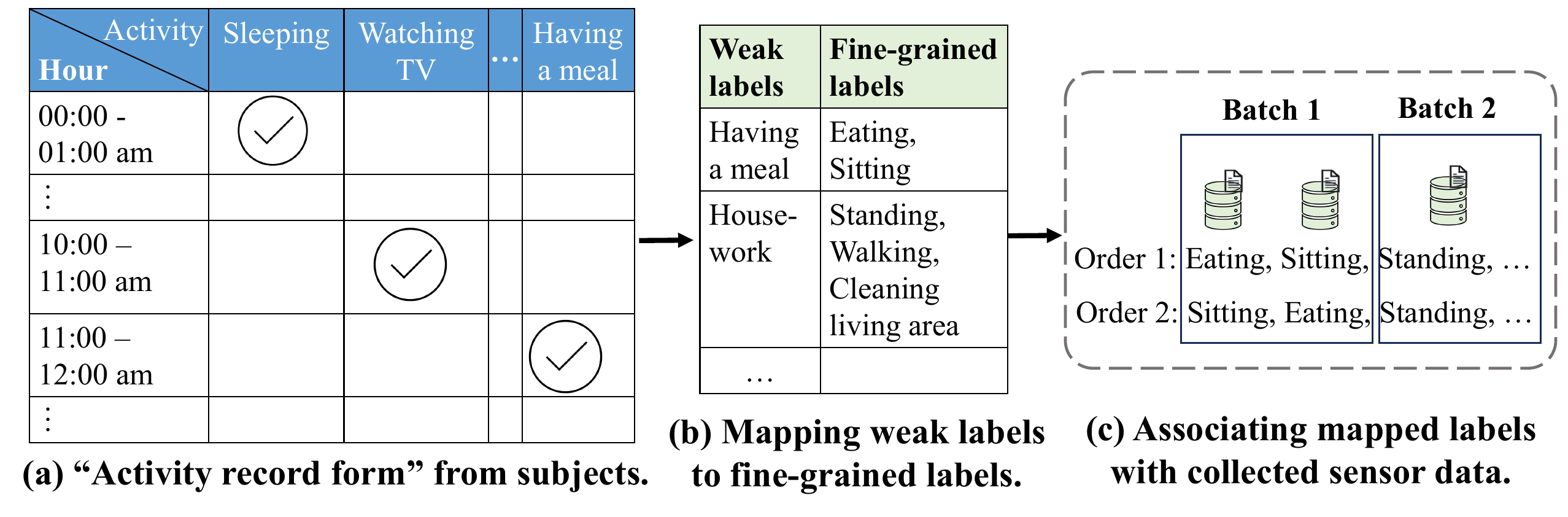}
  \caption{Training with weak labels. The ``activity record form'' provided by subjects is mapped to fine-grained labels and associated with collected data.}
  \label{fig:weak_form}
  \vspace{-1.5em}
\end{figure}

\textbf{Training with weak labels.} 
ADMarker leverages sparse activity logs for weakly supervised training. Such logs can be obtained in several different ways. Many patients are routinely suggested by the doctors to keep a journal of activity logs through ADL scales \cite{patterson1992assessment}. Alternatively, the patients and/or their caregivers may be asked to keep an activity log for the purpose of user training during the initial phase of system deployment. Our results show that ADMarker achieves good performance even if activity logs are available for only several days (see Figure \ref{fig:data-summary_num}). Moreover, only major daily routines like ``sleeping'', ``having a meal'' (see Fig. \ref{fig:weak_form}(a)) are needed, which alleviates the burden of users. 


To train the model with weak labels, we first need to map them to a series of fine-grained activity labels we are interested in (see Table \ref{table:task_index}), which requires sophisticated domain knowledge. For example, the weak label ``Having a meal'' corresponds to the fine-grained labels ``eating'' and ``setting'', while ``Household'' includes the activities of ``standing'' and ``walking''. Another challenge is to associate the mapped fine-grained labels with collected sensor data, because the mapping from weak to fine-grained labels does not include the order of the events. Therefore, during the weakly supervised training, we split the training data such that the data samples in the same batch are collected successively in the time order. Then, to improve the training performance, we shuffle the training samples in each batch to simulate different orders of the activities, and calculate the cross entropy loss with different label permutations for model training.







 \begin{figure*}
     \setlength{\abovecaptionskip}{0.cm}
    \setlength{\belowcaptionskip}{0.cm}
    \centering
  \begin{subfigure}{.33\linewidth}
    \setlength{\abovecaptionskip}{0.cm}
    \setlength{\belowcaptionskip}{-0.cm}
    \centering
     \includegraphics[width = \textwidth]{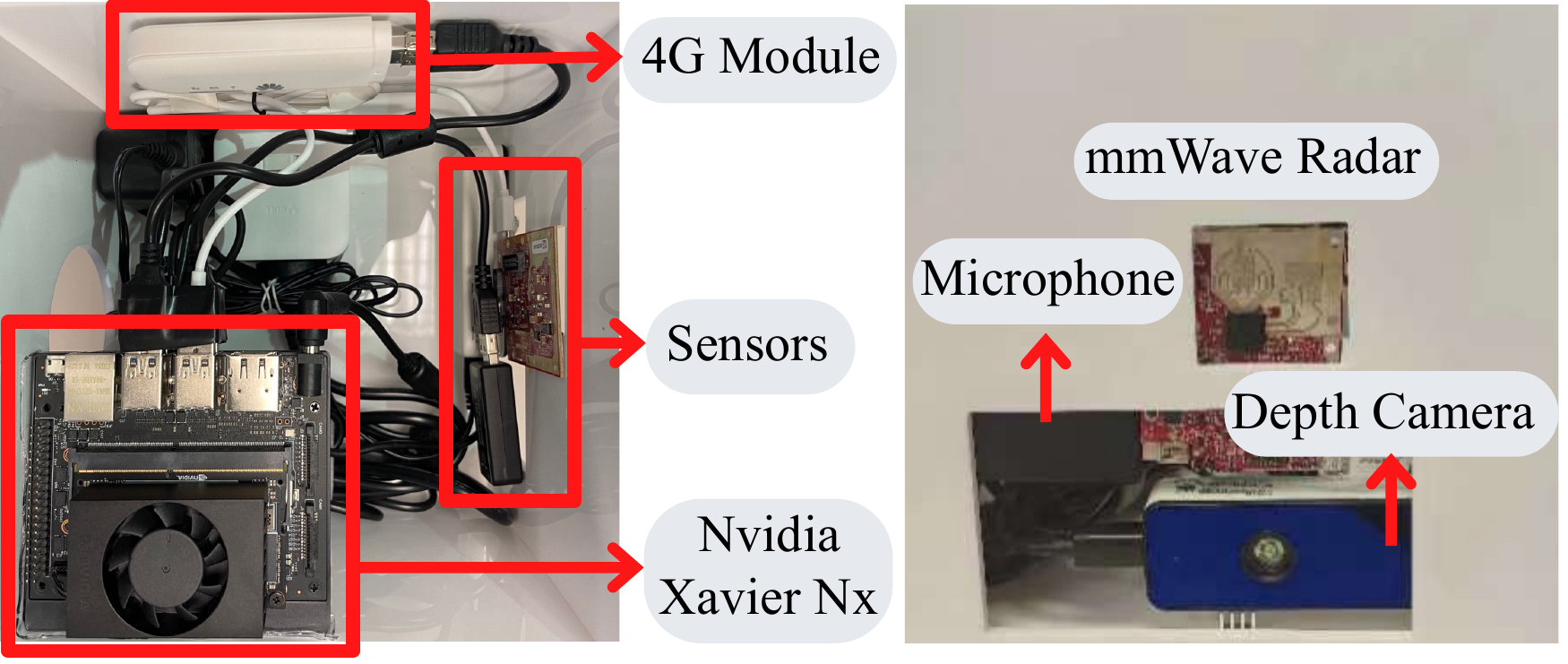}
    \caption{Components and Layout.}
      \label{fig:ADBox-Components}
  \end{subfigure}
      \begin{subfigure}{.29\linewidth}
    \setlength{\abovecaptionskip}{0.cm}
    \setlength{\belowcaptionskip}{-0.cm}
    \centering
     \includegraphics[width = \textwidth]{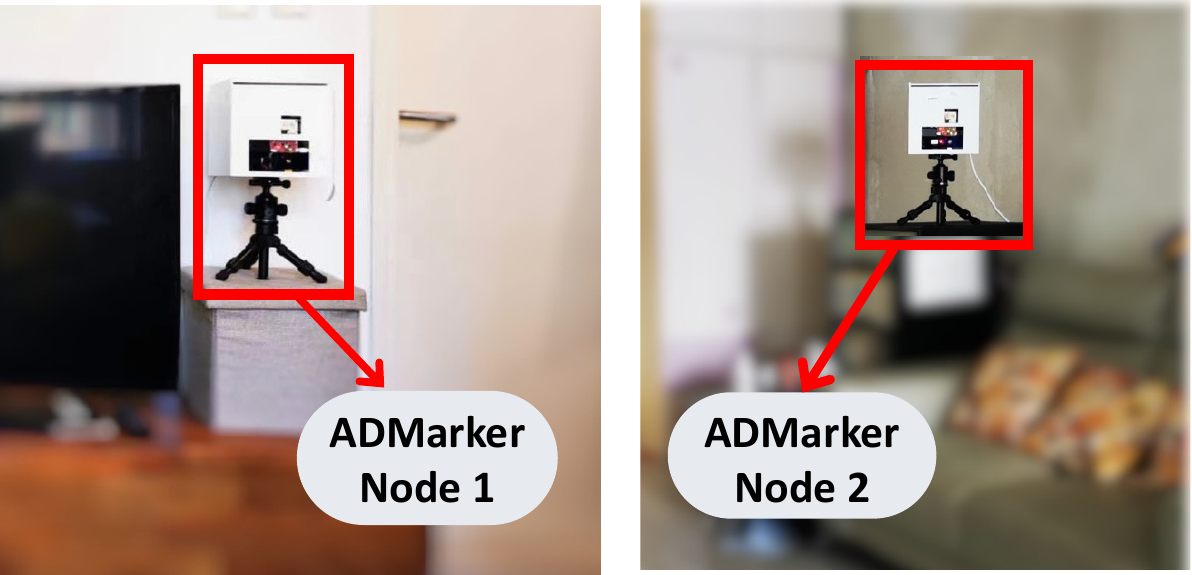}
     \caption{Typical Home Deployments.}
     \label{fig:ADBox-deployments}
    \end{subfigure}
      \begin{subfigure}{.37\linewidth}
    \setlength{\abovecaptionskip}{0.cm}
    \setlength{\belowcaptionskip}{-0.cm}
    \centering
     \includegraphics[width = \textwidth]{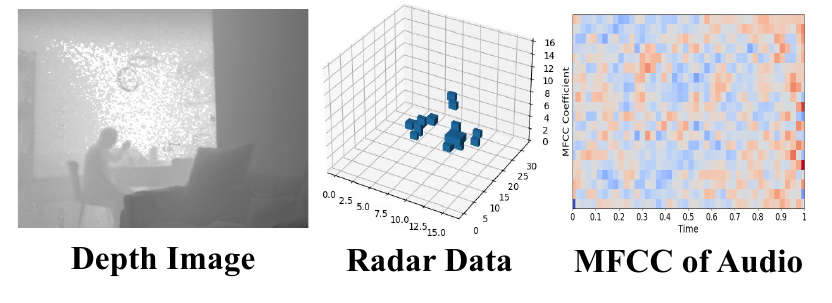}
     \caption{Examples of recorded multi-modal data.}
     \label{fig:ADBox-data}
    \end{subfigure}
    \caption{The ADMarker prototype. The hardware incorporates three sensor modalities (depth, mmWave radar, and audio) to detect multi-dimensional behavior biomarkers in home environments.}
    \label{fig:ADBox_Overview}
    \vspace{-0.5em}
\end{figure*}

\textbf{Local and global balancing on nodes.} To address the challenge of local class imbalance, the nodes will perform both self and global balancing during model training. First, the nodes will apply balanced re-sampling to the train samples during local model training, which avoids the gradient dominance of the majority classes. 
Moreover, as FL proceeds, the server-side global model more or less accumulates some knowledge on all classes \cite{balancefl}. Therefore, we use the aggregated global model as the teacher model to guide the training of the local model using knowledge distillation \cite{gou2021knowledge}.

%% file: 5_medical.tex
\section{Utilizing Digital Biomarkers for AD Analysis}
\label{sec:medical}


 To evaluate the effectiveness of the detected biomarkers, we use a DNN for AD diagnosis, where the inputs are the features of the digital biomarkers and outputs are the predictions of diagnosis results. Then, the major challenge is how to quantify the digital biomarkers to generate effective features as the input of the disease diagnosis model. 
 According to medical literature \cite{dawadi2015automated}, we calculate the duration and frequency of the detected activities over the period of deployment as the features. However, due to the system dynamics (see Section \ref{sec:system_dynamics}), the total duration of the recorded sensor data may vary among different subjects. For example, most of the subjects will have four-week data, while some subjects only have one-week or two-week data due to sensor faults. Therefore, we further normalize the duration and frequency of detected activities with the period of the collected data. 

Moreover, to further select more effective digital biomarkers, we use one-way Analysis of Variance (ANOVA) \cite{st1989analysis} to measure the correlations between each digital marker and the diagnosis results. Before applying ANOVA analysis to the data collected in our clinical deployment, we need to ensure that our data satisfies the following three conditions: independence, normality, and equality. First, the selected features of each individual are independent of others, either in the same or different subject groups (NC, MCI, and AD). Second, we apply the Box-Cox data transform \cite{sakia1992box} to the selected features, ensuring that the transformed features satisfy the normal distributions.
Third, we use the Levene's test \cite{brown1974robust} to access the variances of each feature among the AD, MCI, and NC groups. The mean p-value (0.659) is larger than 0.05, which shows the equality of selected features.

%% file: 6_implementation.tex
\section{System Implementation}
\label{sec:implementation}

\subsection{Hardware System}

Figure~\ref{fig:ADBox_Overview} shows the overview of our ADMarker prototype. 

\subsubsection{Hardware choices.} The goal of the hardware design is to capture lots of digital biomarkers in a privacy-preserving manner while ensuring the durability and scalability of the system. First, {we choose three sensor modalities: a depth camera, an mmWave radar, and a microphone, which collectively capture a wide range of biomarkers while preserving users’ privacy. In particular, the Time-of-Flight (ToF) depth camera can detect context-aware activities like cleaning living areas and moving in/out of chairs, without revealing sensitive personal information like faces. The microphones can help distinguish acoustic-related activities like watching TV and talking with others, which run real-time algorithms to extract Mel-frequency cepstral coefficient (MFCC) features \cite{ittichaichareon2012speech} without storing raw acoustic data. 
The mmWave radar can capture motion-related activities like walking, standing, and sleeping.}
Second, we choose the NVIDIA Xavier NX \cite{xavier-nx} as the main compute unit as it incorporates powerful NVIDIA GPUs (384-core Volta) and CPUs (6-core @1.9GHZ) for on-device model training. Third, to improve the durability of the system, we choose the NVMe SSD rather than the conventional portable HDD or SSD as the external data storage unit. The reason is that they have a relatively lower read/write speed (about 100MB/s), poor reliability (vulnerable to vibration and power failure), and a larger size/weight, which is unsuitable for long-term operation and mass deployment.



\subsubsection{Layout design.} The design of our hardware system comprises a number of components (e.g., sensors, accessories, and cables) in a single box, which increases the difficulty of box assembly and heat dissipation. We carefully optimize the cabling and group similar components within the same shelve, making the box compact and lightweight. The size of the hardware box is about 20cm x 20cm x 20cm. Moreover, the sensing coverage of the sensors is limited, e.g., with the range of 0.35m-4.4m and a field-of-view (FOV) of 69°(H) x 51°(V) for the depth camera. In order to capture the main area of a living room, we add a tripod at the bottom of the box to adjust the height and angle of the box.

 \subsection{Software System}
 
We now present the design of several major functions to improve the stability and scalability of the software system.

\subsubsection{Long-term multi-modal sensor data recording and pre-processing} \label{sec:data_process} 
First, ADMarker needs to collect and store the data of multiple sensors continuously for up to months, which requires a large storage space and incurs significant power consumption. In particular, we set the sampling rates of the depth camera, mmWave radar, and microphone as 15 Hz, 20 Hz, and 44,100 Hz, respectively, which will result in around 4TB data during a four-week deployment.
To address this challenge, we save the depth data in an 8-bit format and compress the images into videos with OpenCV MJPG \cite{opencv-mjpg}, bringing about a 75\% reduction in data volume without significantly sacrificing the data quality. Second, the sensors may stop data recording occasionally, e.g., due to power surges or unstable sensor connections. We use the \emph{systemd} service \cite{systemd} of Linux to restart the sensors in case of sensor failures. Finally, to train the multi-modal models, the recorded sensor data is split into 2-second samples, and then converted into a fixed dimension [16,112,112], [20,2,16,32,16], and [20,87] for depth (cropped images), radar (voxels), and audio (Mel-frequency cepstral coefficients), respectively.

\subsubsection{Improving training and inference efficiency} \label{sec:data-selection} \label{sec:end-to-end-inference}
During online FL, a major challenge of continuous training and inference with all collected sensor data is the significant delay. For example, Table \ref{table:performance_setting} shows that unsupervised FL with 4,000 samples (i.e., data collected in about 2.2 hours) will incur a training delay of 85 hours (3-4 days). This will not only increase the data storage requirement on the device, but also affect the model accuracy due to the delayed updating of collected data during model training. Therefore, we propose two approaches to improve training and inference efficiency.

First, we apply the following online data selection strategies to reduce the model training delay while maintaining the effectiveness of AD symptom monitoring. Basically, the sensors will only collect data during a 12-hour period (i.e., 7:00-19:00) that contains fundamental daily activities, such as having meals, house-holding, etc. Moreover, rather than saving all recorded data, we evenly sample the data over time and only choose 1\% of the data for model training. The reason is that most activities of interest, such as sitting, walking, or standing, last for a relatively long time (e.g., several minutes). 
Finally, we use Yolov5 \cite{yolov5} to detect humans in the depth images and delete the data samples without humans. 

\begin{figure}
    \setlength{\abovecaptionskip}{0.cm}
    \setlength{\belowcaptionskip}{0.cm}
    \centering
     \includegraphics[width = 0.9\linewidth]{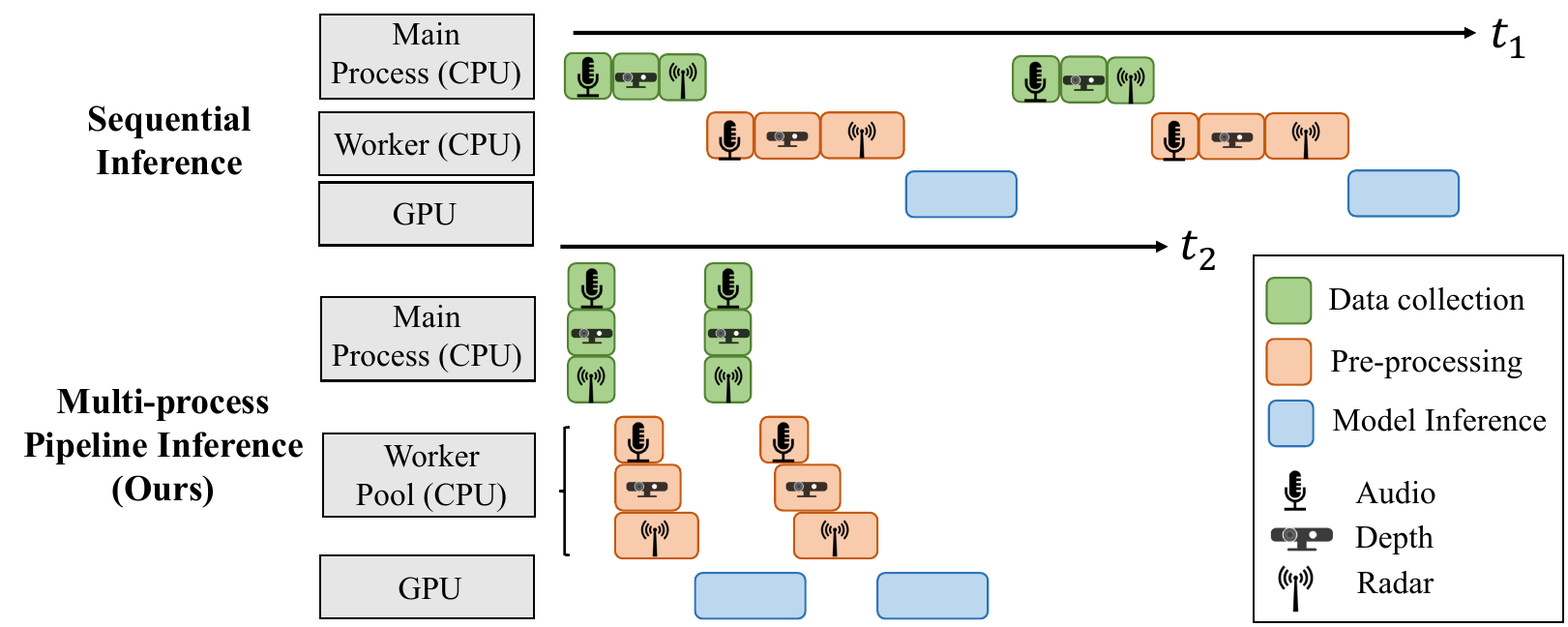}
  \caption{Illustration of our multi-process pipeline inference and conventional sequential inference scheme.}
  \label{fig:pipeline}
  \vspace{-1em}
\end{figure}

To reduce the on-device inference latency, we design a multi-task scheduling scheme for accelerating the pipeline in end-to-end multi-modal inference, including multi-sensor data collection, data pre-processing, and model inference. 
 As shown in Figure \ref{fig:pipeline}, ADMarker will run the three major tasks in parallel instead of processing them sequentially. For example, during the inference of previous data frames on GPU, the tasks of recording and pre-processing the incoming data frames are scheduled to utilize the spare resources on CPUs.
 Moreover, ADMarker maintains a pool of processes to handle data pre-possessing tasks of different modalities in parallel. As a result, ADMarker can detect biomarkers in real-time, e.g., 9.45 frames per second (of depth data) in the end-to-end multi-modal inference.

\subsubsection{Private and stable communication networks} \label{sec:comminucation}
To enable federated learning, the system should have highly stable and secure Internet connectivity to the server. However, the home WiFi or Ethernet connection of users should not be used due to privacy and cost concerns. Therefore, the ADMarker nodes are incorporated with a cellular interface to communicate with the server using 4G LTE through a Virtual Private Network (VPN). Moreover, to reduce the communication delay, we dynamically select the optimal frequency band on each node according to the runtime demand of FL, e.g., B3 (1800Mhz, 21Mbps) when uploading models to the server and B40 (2300Mhz, 20Mbps) when downloading models.

%% file: 7_deployment.tex
\section{Clinical Deployment}
\label{sec:deployment}


A total of 91 elderly subjects (43 females and 48 males), aged between 61 and 93 years old, have participated in our clinical study\footnote{All the data collection was approved by IRB and the Clinical Research Ethics Committee of the authors' institution.}. As shown in Table \ref{table:sunjetc_statistics}, the participants were from three groups: 31 with Alzheimer's Disease, 30 with mild cognitive impairment (MCI), and 30 are cognitively normal. The ADMarker node will be installed at the height of 1.5m-1.8m in the living room of the subject's home for four weeks, as shown in Figure~\ref{fig:ADBox-deployments}. {We only installed one node in the living room, not only because this is the area where most activities happen but also brings less concerns of privacy. During deployment, we used a tripod to adjust the angle of the hardware box to cover the main living area of the subject.} The installation process typically takes about ten minutes per home. The subjects and their caregivers were asked to fill in an ``activity record form'' by ticking the relevant daily activities, which served as the weak labels for behavior analysis (see Section \ref{sec:weak_supervise_fl}).

\begin{table}
    \centering
 \setlength{\abovecaptionskip}{0.cm}
    \setlength{\belowcaptionskip}{0.cm}
    \resizebox{\linewidth}{!}{
     \begin{tabular}{cccc}
      \toprule
      Group & AD & MCI & NC\\
      \midrule
      No. of subjects & 31 (M/F:10/21) & 30 (M/F:18/12) & 30 (M/F:20/10)\\
      Age (Mean$\pm$STD) & 79.53$\pm$7.11 & 78.14$\pm$5.73 & 70.66$\pm$6.94 \\
      Living Mode (A/W-F/W-C) & 2/25/4 & 2/27/1 & 6/24/0 \\
      Education Years (Mean$\pm$STD) & 6.16$\pm$3.82 & 
      5.79$\pm$3.70 & 
      10.59$\pm$4.10 \\
      MoCA Score (Mean$\pm$STD) & 11.75$\pm$6.38 & 19.50$\pm$6.53 & 25.07$\pm$4.41 \\
      \bottomrule
    \end{tabular}}
    \caption{Demographic characteristics of enrolled elderly subjects (N = 91). A: Alone; W-F: With Family; W-C: With Caregiver. MoCA: Montreal Cognitive Assessment \cite{nasreddine2005montreal}.}
    \label{table:sunjetc_statistics}
    \vspace{-2em}
\end{table}

\textbf{Diagnosis of the subjects.} 
A neuropsychologist in our study gives diagnosis results of the subjects, which will be used to enroll subjects with a balanced distribution and evaluate the performance of our detected digital biomarkers.
Before the study, each subject underwent a screening test (i.e., Montreal Cognitive Assessment \cite{nasreddine2005montreal}, MoCA) to generate a cognitive score. A low MoCA score indicates the risk of cognitive impairment.
The subjects then received physical examinations and face-to-face consultations with a neuropsychologist if they: (1) have no medical record of AD diagnosis but receive low MoCA scores; (2) already have a medical record of AD diagnosis before, but their cognitive scores are obviously inconsistent with the medical records. Moreover, patients with AD and MCI who have not received an MRI test within one year are suggested to do a free MRI test, while the MRI test is voluntary for cognitively normal subjects. Then, the neuropsychologist will give the final diagnosis results of the enrolled subjects by combining the results of the face-to-face interview and MRI scan. 

%% file: 8_experiments.tex
\section{Evaluation}
\label{sec:evaluation}

\subsection{Methodology}

\textbf{Experiment settings.} Our evaluation is based on the clinical study of total 91 subjects. The ADMarker systems deployed in these subjects' homes collected a total of 61,152 hours of multi-modal sensor data, with a total size of over 91TB. Among these, the data of 31 subjects is used for model pre-training, and 60 deployed nodes run the unsupervised and weakly supervised federated learning algorithms continuously for four weeks. 
{We analyze the data from all 91 subjects to show the characteristics of data distributions in Section \ref{sec:data_overview}. However, we only use data from 69 subjects (22 AD, 25 MCI, and 22 NC) for evaluation, because the remaining 22 subjects either have less than one week of valid sensor data or limited labeled data that contains humans (e.g., tens of samples), which would cause highly biased results.
To evaluate the performance of AD diagnosis using the detected digital biomarkers, we adopt 3-fold cross-validation using the data of the 69 subjects. }
We use the software packages \emph{jetson-stats}  from JetPack 4.6.1 \cite{jetson-stats} to record the power, memory/CPU/GPU usage and temperature of the hardware system, and use \emph{speedtest-cli} \cite{speedtest-cli} to measure the network bandwidth and latency.



\textbf{Configurations of models.} {The multimodal activity recognition model contains a 5-layer 2D-CNN for audio MFCC features, an 8-layer 3D-CNN for depth, a 5-layer 3D-CNN for mmWave radar, plus 3-layer multi-layer perception (MLP) that is input with concatenated unimodal features for activity recognition.} The learning rate and batch size are 0.01 and 16 for unsupervised FL, and 0.001 and 16 for supervised FL. We use the data collected in the first and last weeks for model training and evaluation, respectively.

\textbf{Data annotation} As a long-term autonomous monitoring system, ADMarker is expected to rely on as few manual annotations as possible. We use the ``activity record form'' (see Section \ref{sec:weak_supervise_fl}) provided by the subjects to generate the weak labels for weakly supervised FL. Moreover, the multi-modal data collected by the system are synchronized using the system clock and annotated using depth videos by a professional data labeling company. 
The manually labeled data is used as ground truth for evaluating the performance of our system and helps to understand the performance of leveraging weak labels.


\begin{figure}
    \setlength{\abovecaptionskip}{0.cm}
    \setlength{\belowcaptionskip}{0.cm}
    \centering
      \begin{subfigure}{0.52\linewidth}
    \setlength{\abovecaptionskip}{0.cm}
    \setlength{\belowcaptionskip}{-0.cm}
    \centering
     \includegraphics[width = \textwidth]{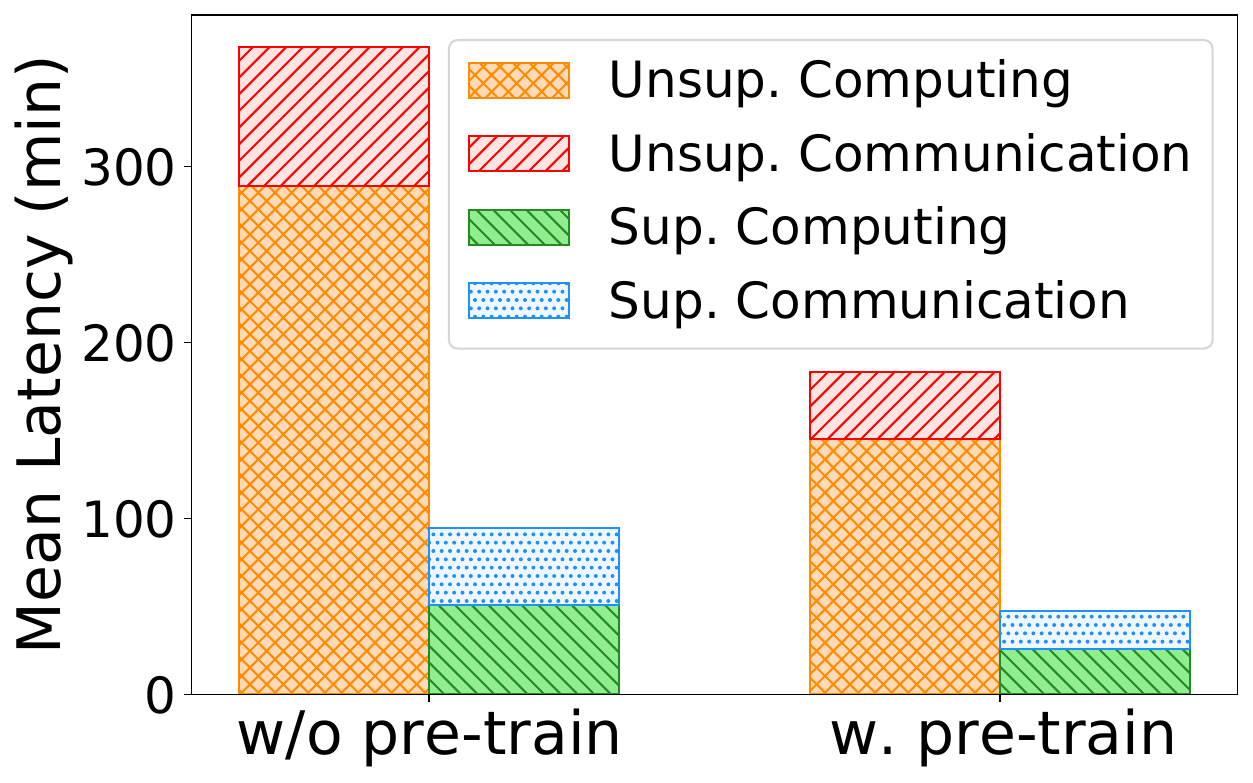}
     \caption{Training latency.}
     \label{fig:fl-overhead-latency_compare}
    \end{subfigure}\hspace{5pt}
      \begin{subfigure}{.42\linewidth}
    \setlength{\abovecaptionskip}{0.cm}
    \setlength{\belowcaptionskip}{-0.cm}
    \centering
     \includegraphics[width = \textwidth]{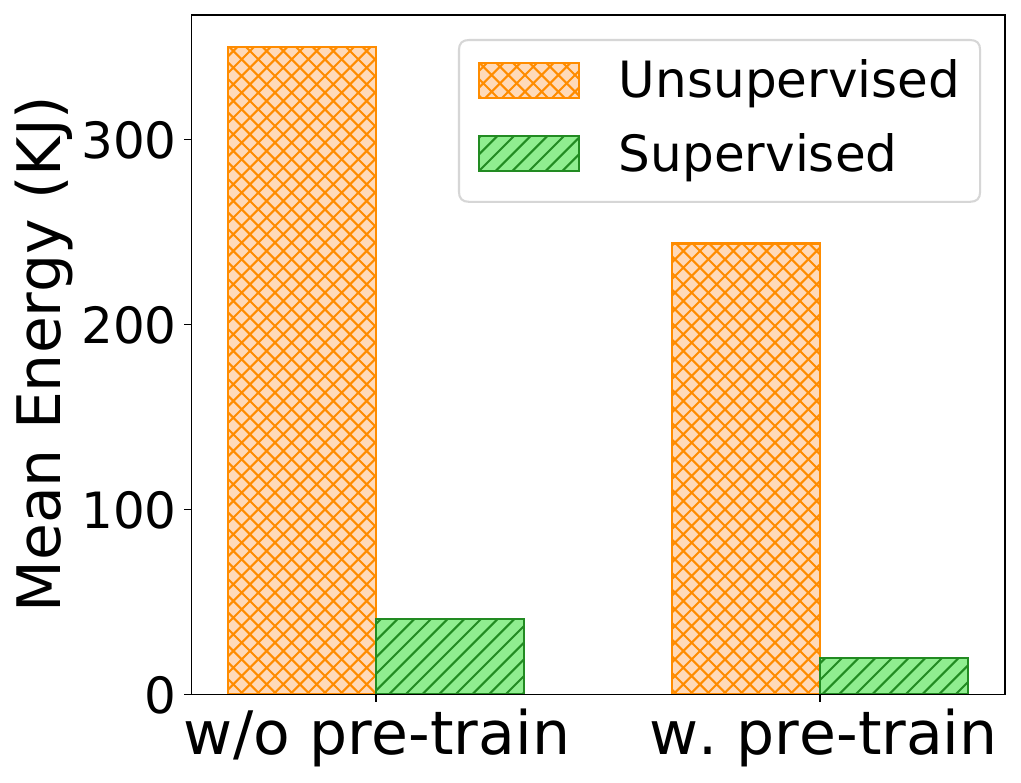}
     \caption{Energy consumption.}
     \label{fig:fl-overhead-power_compare}
    \end{subfigure}
  \caption{System overhead with or without the pre-trained model. ``Unsup. Computing'' denotes the computing time of nodes in unsupervised FL.}
  \label{fig:system-overhead}
  \vspace{-1.em}
\end{figure}

\begin{figure}
    \setlength{\abovecaptionskip}{0.cm}
    \setlength{\belowcaptionskip}{0.cm}
    \centering
     \includegraphics[width = \linewidth]{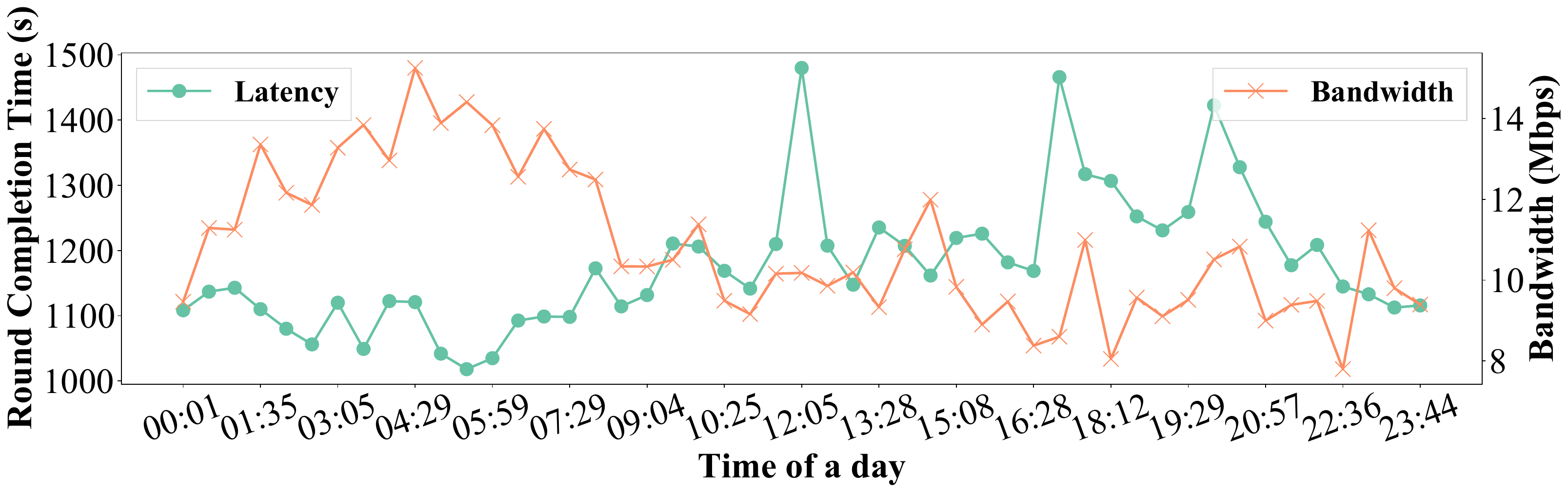}
  \caption{Round completion time of FL and mean bandwidth in a 24-hour operation.}
  \label{fig:fl-overhead-round_time}
  \vspace{-1.5em}
\end{figure}

\subsection{System Overhead}
In this section, we evaluate the system overhead of ADMaker during federated learning in the presence of various system dynamics (see Section \ref{sec:system_dynamics}). 
For example, during the 4-week deployment, the sensors may stop data recording occasionally due to power surges or unstable sensor connections. The bandwidth of nodes also fluctuates over time. 


Figure~\ref{fig:system-overhead} compares the mean training latency and energy consumption with or without centralized model pre-training in online FL. Here, the numbers of training samples in unsupervised and unsupervised FL are 300 and 100, respectively. When the models of nodes are initialized with the pre-trained model, the system overhead of both unsupervised and unsupervised FL is significantly reduced. For example, in unsupervised FL, the overall training latency and the energy consumption are reduced by 184.3min and 106KJ, respectively, since the models converge faster than training from scratch. Moreover, the system overhead of supervised FL is much smaller than unsupervised FL, because the feature encoder networks are already trained with large amounts of unlabeled multi-modal data in unsupervised FL.

Figure~\ref{fig:fl-overhead-round_time} plots the variation of round completion time of FL and mean downlink bandwidth of nodes at different time of the day. The round completion time denotes the latency of finishing one global round of unsupervised FL (i.e., computing time for local training plus communication time). Generally, the training latency of one FL round at nighttime is shorter than at daytime (by about 14\%), because of a better 4G LTE network connectivity. Moreover, the latency increases significantly at around 12:00, 17:00, and 19:00, which is probably caused by the increased network traffic from smartphone users during/after mealtimes. Therefore, the nodes in ADMarker could run federated learning at nighttime to reduce the overall training latency. We also note that the network congestion or node disconnectivity will not affect the accuracy of biomarker detection. 

\begin{figure}
     \setlength{\abovecaptionskip}{-0.cm}
    \setlength{\belowcaptionskip}{0.cm}
    \centering
     \includegraphics[width = \linewidth]{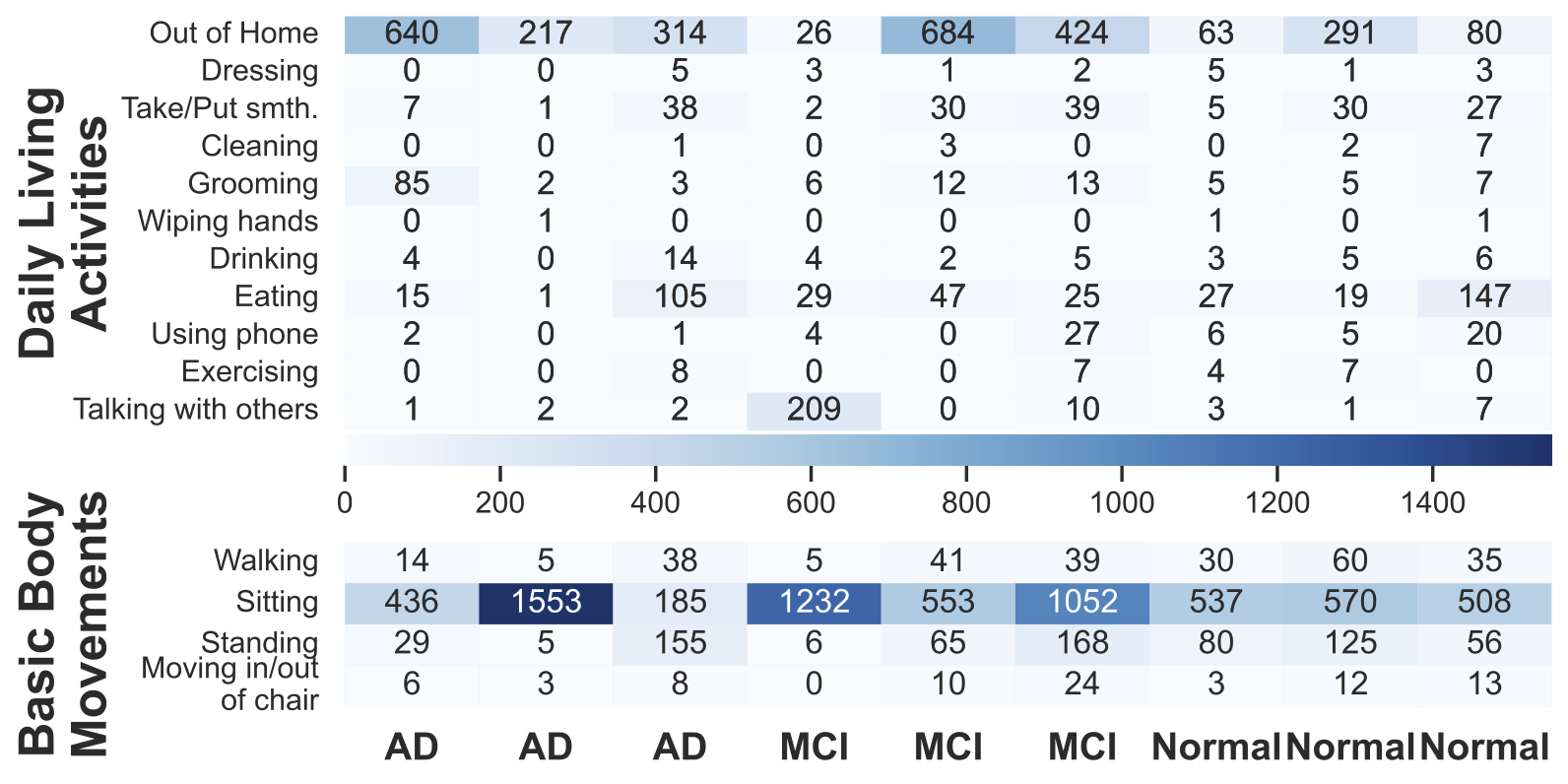}
     \caption{Examples of activity distribution across nine subjects. The numbers denote the amount of data samples of the activities over four weeks.}
     \label{fig:data-overview-distribution}
     \vspace{-1em}
\end{figure}

\begin{figure}
    \setlength{\abovecaptionskip}{0.cm}
    \setlength{\belowcaptionskip}{0.cm}
    \centering
      \begin{subfigure}{.5\linewidth}
    \setlength{\abovecaptionskip}{0.cm}
    \setlength{\belowcaptionskip}{-0.cm}
    \centering
     \includegraphics[width = \textwidth]{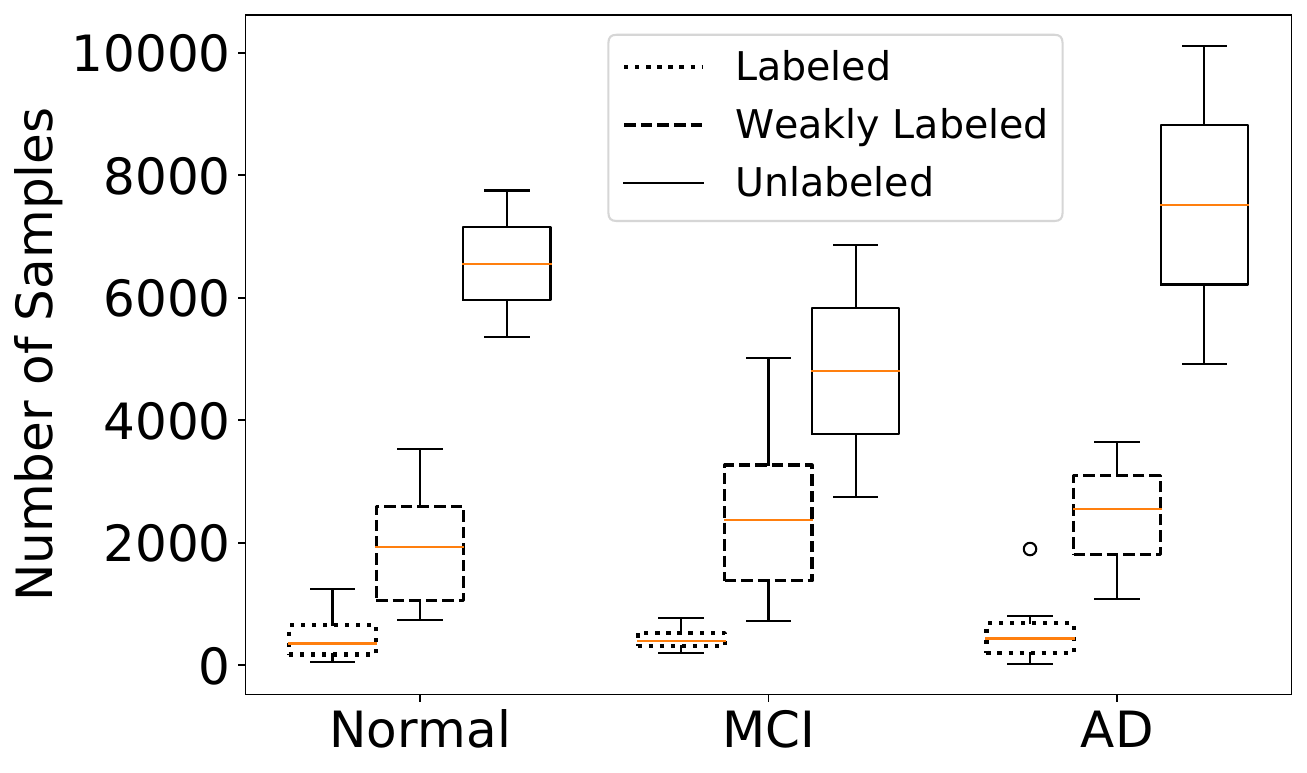}
     \caption{Amount of data.}
     \label{fig:data-summary_num}
    \end{subfigure}
      \begin{subfigure}{.47\linewidth}
    \setlength{\abovecaptionskip}{0.cm}
    \setlength{\belowcaptionskip}{-0.cm}
    \centering
     \includegraphics[width = 0.9\textwidth]{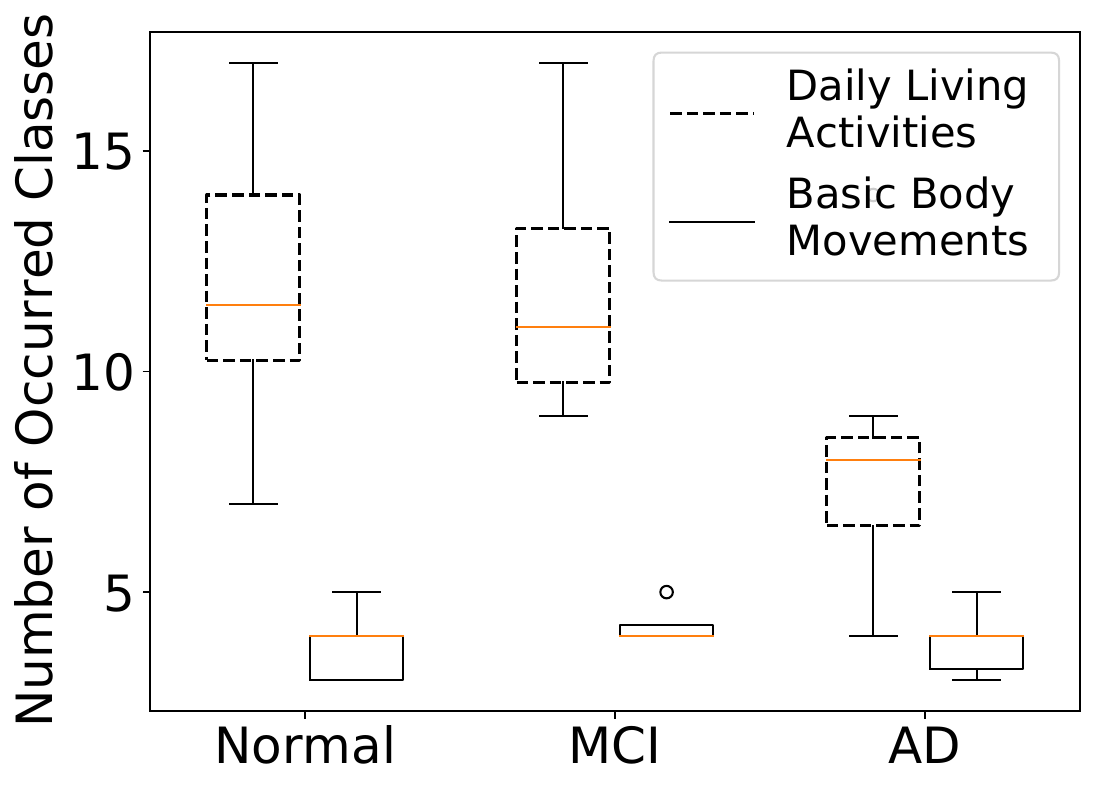}
     \caption{Diversity of activities.}
     \label{fig:data-summary_class}
    \end{subfigure}
  \caption{Data summary of different subject groups.}
  \label{fig:data-summary}
   \vspace{-1em}
\end{figure}

\subsection{Accuracy of Biomarker Detection}

\subsubsection{Overview of the subjects' data} \label{sec:data_overview}We first show the overall characteristics of data collected by ADMarker during the four weeks of real-world deployment. First, Figure \ref{fig:data-overview-distribution} shows examples of the class distribution across nine subjects for daily living activities and basic body movements, respectively. We can observe obvious class imbalance in different subjects' local data. Moreover, the class distributions of different subjects are highly non-i.i.d. 
Second, Figure~\ref{fig:data-summary_num} shows that there is only limited labeled sensor data during real-world deployment, due to the significant overhead and cost of data annotation. For example, the average amount of labeled data is less than 10\% of unlabeled data. However, the amount of data with weak labels (see Section \ref{sec:weak_supervise_fl}) is much larger than manually annotated data, which can be leveraged to effectively train the models.
Finally, Figure~\ref{fig:data-summary_class} presents the number of occurred activities during the deployment. In daily living activities, cognitively normal and MCI subjects exhibit more diverse activities than AD subjects, which shows the decline in cognitive and functional ability during the progression of AD \cite{livingston2017dementia}.

\begin{figure}
     \setlength{\abovecaptionskip}{-0.cm}
    \setlength{\belowcaptionskip}{0.cm}
    \centering
     \includegraphics[width = \linewidth]{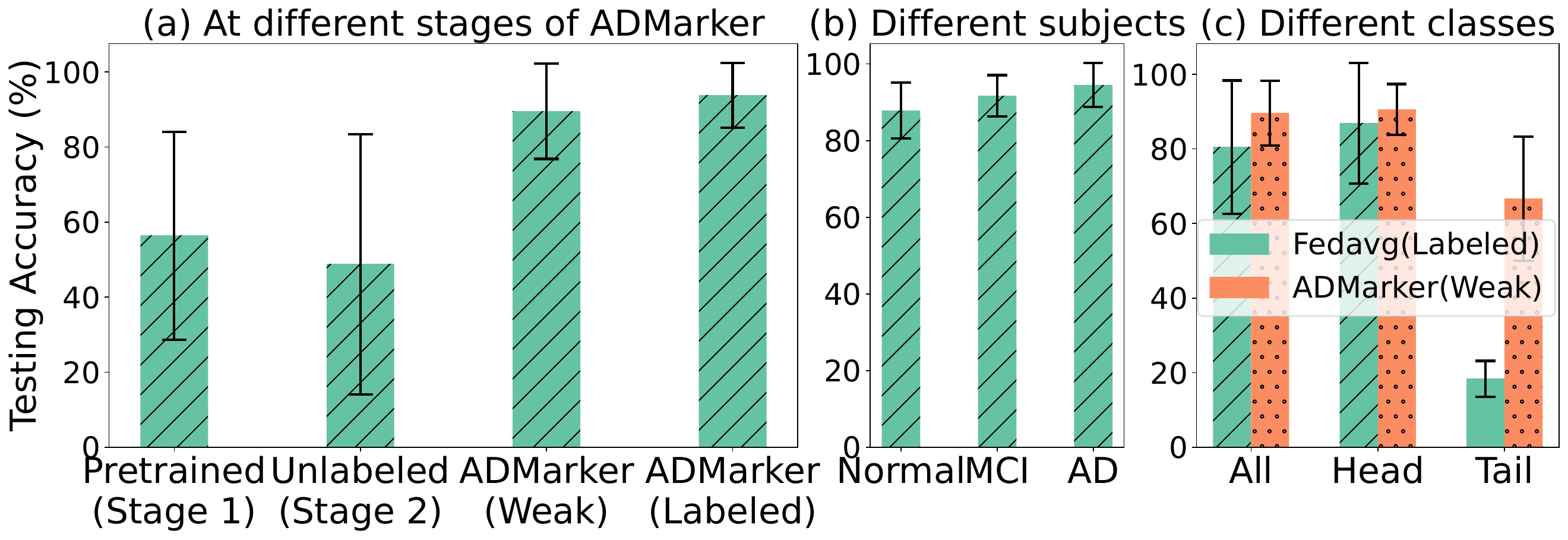}
     \caption{Accuracy of biomarker detection at different stages, on different subject groups and activity classes.}
     \label{fig:task-accuracy}
     \vspace{-0.5em}
\end{figure}

\subsubsection{Performance of biomarker detection} Figure~\ref{fig:task-accuracy} shows the accuracy of biomarker detection under different settings.
\label{sec:accuracy_biomarker_detection}

\vspace{-1em}
\textbf{Performance at different stages of FL.}
Figure~\ref{fig:task-accuracy}(a) shows the accuracy at different stages of federated learning. First, directly applying the pre-trained model has a very low detection accuracy on the participants during the deployment due to the large domain gap, e.g., with a mean accuracy of 56.37\%. 
Second, although the model accuracy after unsupervised FL (Stage 2 of ADMarker) is still low since the classifier layers are not trained, the feature encoders trained using large amounts of unlabeled data can significantly improve the accuracy performance at Stage 3. Finally, full-fledged ADMarker with a three-stage FL design can improve the model accuracy by combining the unlabeled and labeled data, e.g., 89.56\% and 93.81\% mean accuracy with weak and annotated labels, respectively.


\textbf{Performance on different subjects.} Figure~\ref{fig:task-accuracy}(b) shows the accuracy of biomarker detection on different subject groups, which varies among the three subject groups.
In particular, the mean accuracy is over 94.51\% for AD subjects, while 87.83\% and 91.67\% for cognitively normal and MCI subjects, respectively. The reason is that the cognitively normal and MCI subjects generally exhibit significantly more diverse behaviors in daily living (see Figure~\ref{fig:data-summary_class}), increasing the difficulty of detection. 

\textbf{Dealing with class imbalance.} Figure~\ref{fig:task-accuracy}(c) compares the performance of Fedavg with manually labeled data and ADMarker with weak labels on detecting different classes of activities. The models are initialized with weights trained at the second stage of ADMarker.
We define the behaviors with the most and least four data samples as the head classes (i.e., walking, sitting, standing, and eating) and tail classes (i.e., cleaning the living area, grooming, wiping hands, and exercising), respectively. Compared with Fedavg, the overall accuracy (all classes) is improved by $\sim$9\%. However, Fedavg exhibits a very bad accuracy in detecting the tail classes, with $\sim$18.33\% on average. Our approach can improve the performance on tail classes by $\sim$48.33\% compared with Fedavg. 


\begin{figure}
    \setlength{\abovecaptionskip}{0.cm}
    \setlength{\belowcaptionskip}{0.cm}
    \centering
      \begin{subfigure}{.5\linewidth}
    \setlength{\abovecaptionskip}{0.cm}
    \setlength{\belowcaptionskip}{-0.cm}
    \centering
     \includegraphics[width = \textwidth]{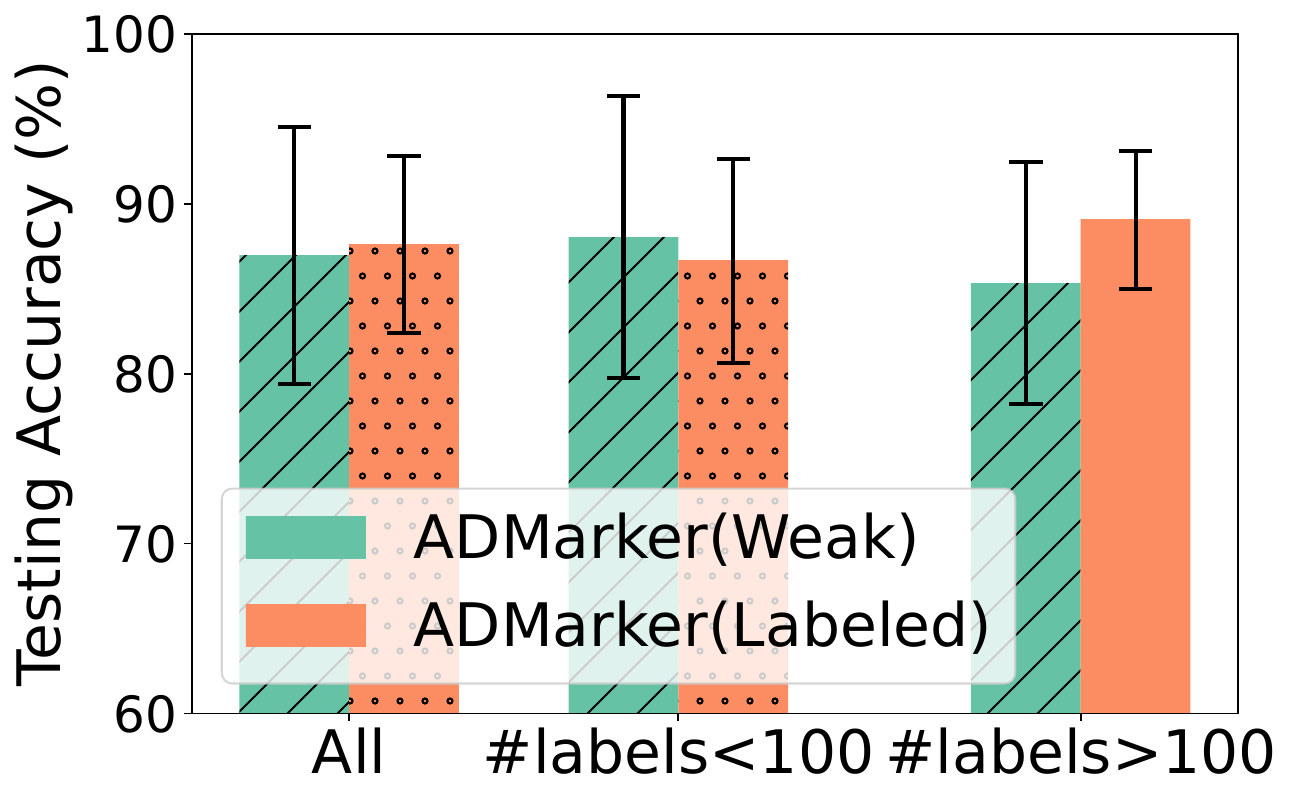}
     \caption{Amount of annotated data.}
     \label{fig:weak_amount}
    \end{subfigure}
      \begin{subfigure}{.48\linewidth}
    \setlength{\abovecaptionskip}{0.cm}
    \setlength{\belowcaptionskip}{-0.cm}
    \centering
     \includegraphics[width = \textwidth]{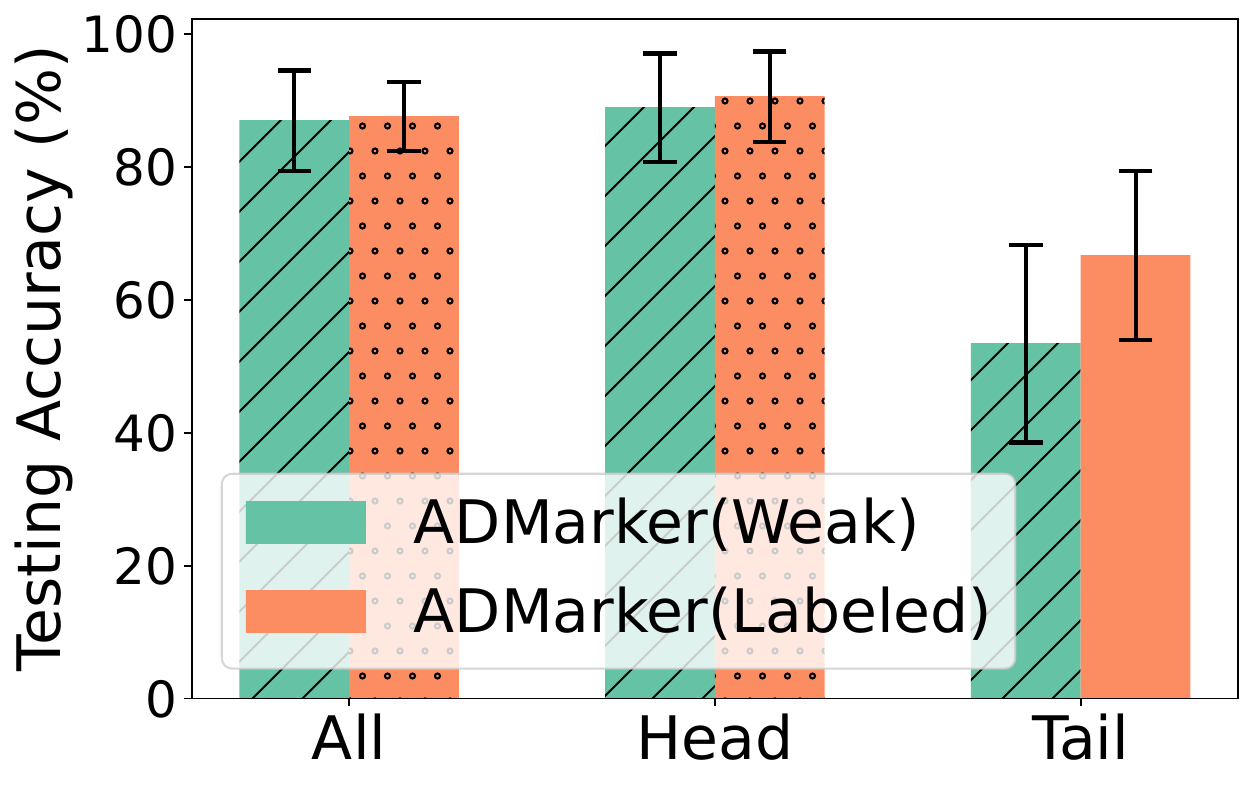}
     \caption{On different classes.}
     \label{fig:weak_class}
    \end{subfigure}
  \caption{Understanding the weak labels by comparing with performance using manually labeled data.}
  \label{fig:weak_labels}
  \vspace{-0.5em}
\end{figure}

\subsubsection{Understanding the weak labels} We further study the effectiveness of learning with weak labels in Section \ref{sec:weak_supervise_fl} by comparing the performance with manually annotated data. To avoid the impact of different subject groups, the results are based on the data of cognitively normal subjects.

\textbf{With different amounts of annotated data.} Figure \ref{fig:weak_amount} compares the performance of two approaches among subjects with different amounts of annotated data. Although the weak labels provided by the subjects are usually noisy and sparse, the amount of associated weakly labeled data is usually larger than manually annotated data (see Figure \ref{fig:data-summary_num}). Therefore, when the subjects have very limited data (\# of labels<100), ADMarker with weak labels even performs better than with annotated labels, e.g., by 1.41\% improvement in mean accuracy. However, when the subjects have enough labeled data (\# of labels<100), ADMarker with annotated data will have a better accuracy performance.

\textbf{On different classes.} We then compare the performance of two approaches on different classes, where the head and tail classes are defined the same as Section \ref{sec:accuracy_biomarker_detection}. Compared with ADMarker with annotated labels, the performance with weak labels is very close on the head classes, while worse on the tail classes. The reason is that the weak labels provided by the subjects are more reliable on  ``persistent'' activities like walking, sitting, and eating, while noisy on dynamic activities like cleaning the living area, wiping hands, and exercising. Therefore, the performance of ADMarker with weak labels can be further improved if we combine only a limited amount of annotated labels on the tail classes.

\subsection{Interpretation of Detected Biomarkers}  
\label{sec:medical_analysis}

\subsubsection{Utilizing the biomarkers for AD diagnosis.} 
\begin{figure}
    \setlength{\abovecaptionskip}{0.cm}
    \setlength{\belowcaptionskip}{0.cm}
    \centering
    \setlength{\abovecaptionskip}{0.cm}
    \setlength{\belowcaptionskip}{-0.cm}
    \centering
     \includegraphics[width = \linewidth]{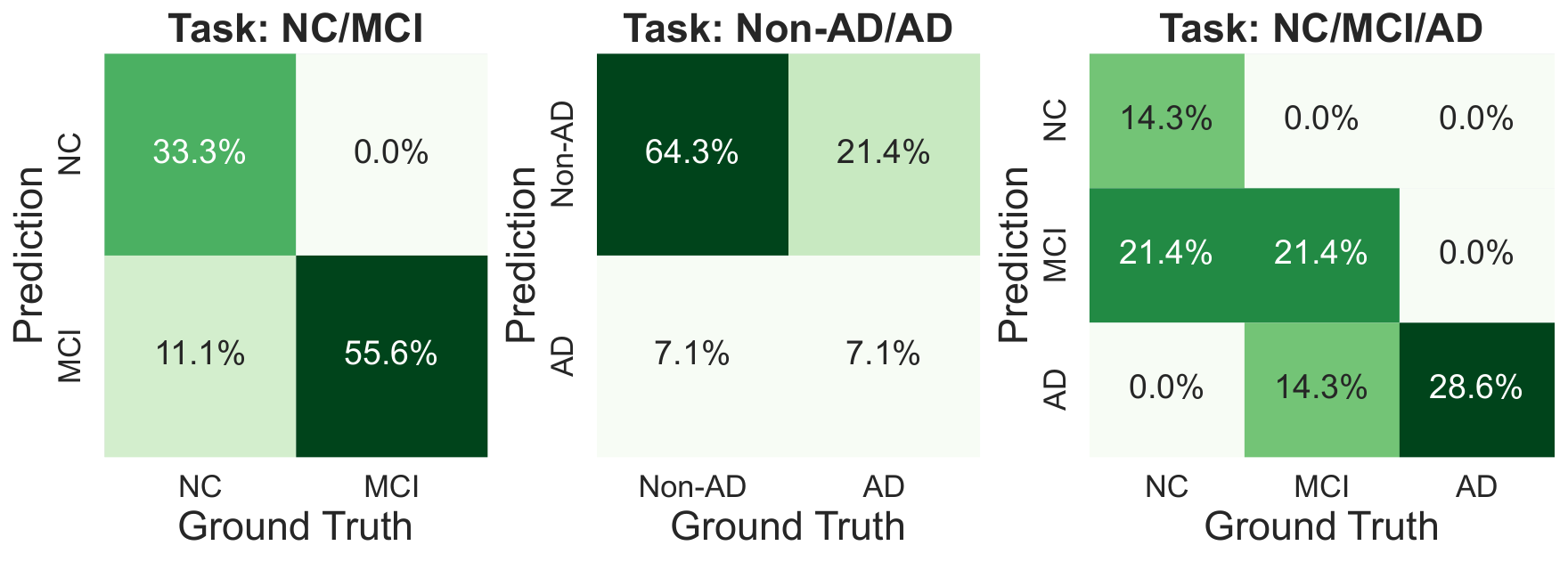}
\caption{The confusion matrix of utilizing the digital biomarkers for different AD diagnosis tasks.}
  \label{fig:heatmap}
  \vspace{-1em}
\end{figure}

Figure~\ref{fig:heatmap} shows the effectiveness of utilizing the detected digital biomarkers for three diagnostic tasks (i.e., normal/MCI, non-AD/AD, and normal/MCI/AD). These tasks are clinically important and are consistent with the current practice of AD diagnosis \cite{marshall2011executive,livingston2020dementia}. 
First, ADMarker achieves about 88.9\% accuracy (33.3\%+55.6\%) in classifying MCI and cognitively normal subjects, meaning that ADMarker is able to identify people at early stage of AD. Second, the accuracy of non-AD/AD and normal/MCI/AD is 71.4\% and 64.3\%, respectively, which shows that identifying MCI from AD subjects is very difficult for subjects with various demographic and medical characteristics. 
Nevertheless, the results are better than the state-of-the-art studies based on a single digital biomarker. For example, Tatc \cite{li2018tatc} identifies MCI with only 42.3\% accuracy, and ADReSS \cite{luz2020alzheimer} achieves 60.8\% for detecting AD. 


\begin{figure}
    \setlength{\abovecaptionskip}{0.cm}
    \setlength{\belowcaptionskip}{0.cm}
    \centering
      \begin{subfigure}{0.48\linewidth}
    \setlength{\abovecaptionskip}{0.cm}
    \setlength{\belowcaptionskip}{-0.cm}
    \centering
     \includegraphics[width = \linewidth]{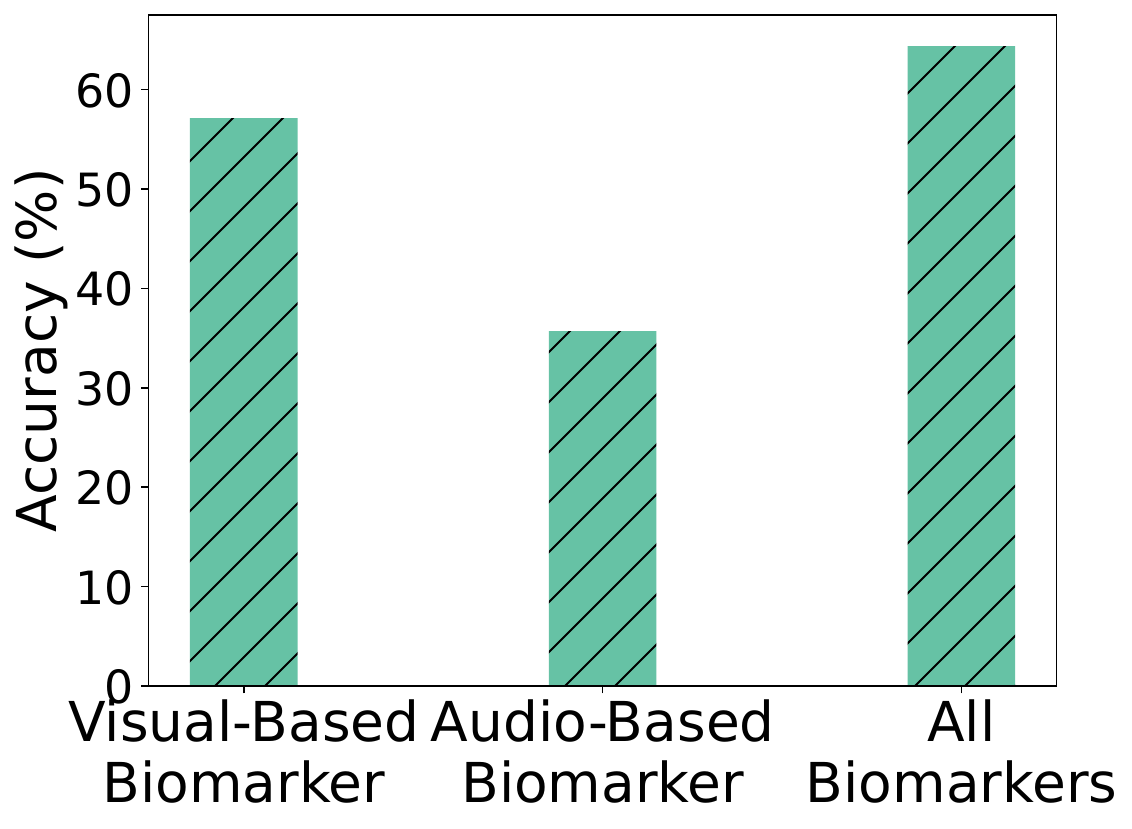}
     \caption{Different modalities.}
     \label{fig:diagnosis_sensors}
    \end{subfigure}\hspace{2pt}
    \begin{subfigure}{0.47\linewidth}
    \setlength{\abovecaptionskip}{-0.cm}
    \setlength{\belowcaptionskip}{0.cm}
    \centering
     \includegraphics[width = \linewidth]{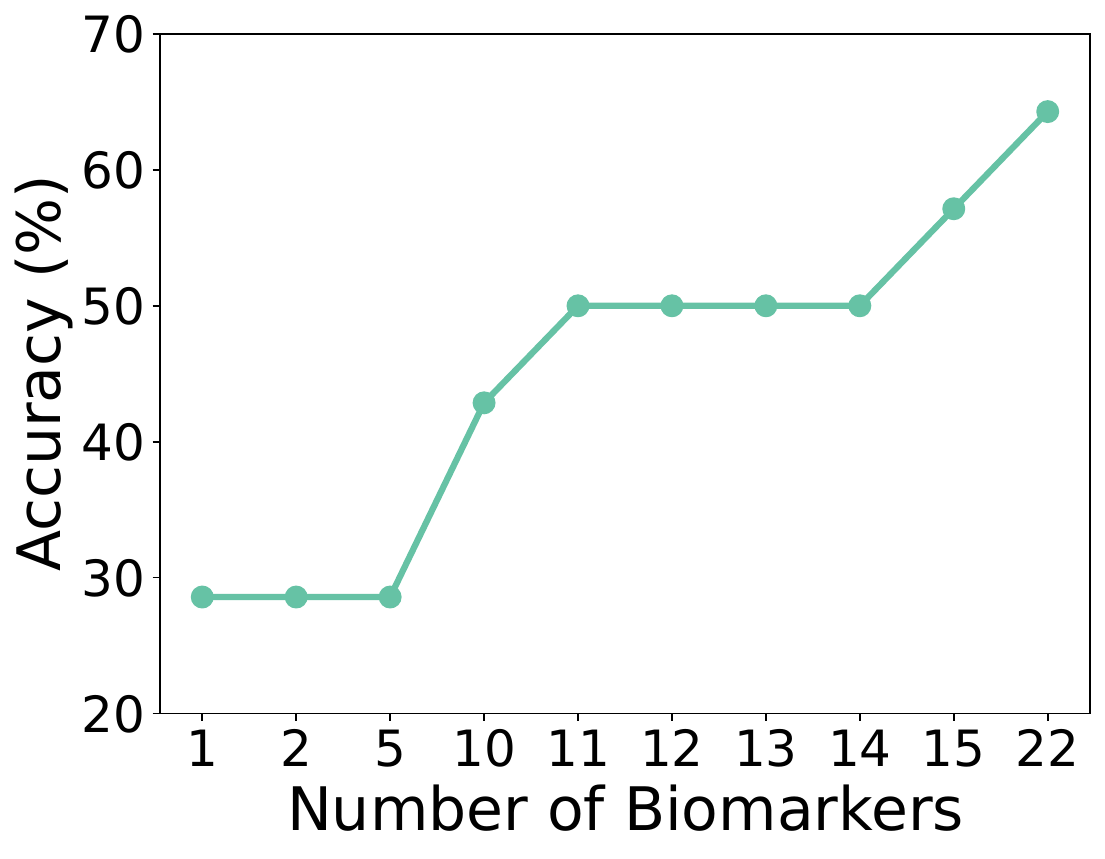}
     \caption{Numbers of biomarkers.}
     \label{fig:diagnosis_num}
    \end{subfigure}
\caption{Impact of leveraging multiple biomarkers.}
  \label{fig:eval_3classes}
  \vspace{-1em}
\end{figure}

\subsubsection{Advantages of leveraging multiple biomarkers.} Figure~\ref{fig:diagnosis_sensors} compares the accuracy of identifying normal/MCI/AD, with vision-based biomarkers (e.g., walking, standing) using depth and radar data, audio-based biomarkers (e.g., talking) that can be captured with microphones, and all biomarkers (e.g., eating, phone call) that need a combination of all three sensors. 
We observe that the diagnosis accuracy with only a single sensor modality is relatively low, e.g., 57.1\% and 35.7\% for vision and audio-based biomarkers, respectively. When multiple sensor modalities are used, the diagnosis accuracy has more than about 30\% improvement over single
modality-based biomarkers. Figure~\ref{fig:diagnosis_num} shows the diagnosis accuracy with different numbers of biomarkers, where the biomarkers are sequentially added with the order of class index in Table~\ref{table:task_index}. The results show that more than a few biomarkers are needed for accurate AD diagnosis, which demonstrates the significance of integrating multiple sensors for detecting multidimensional biomarkers in ADMarker.

\subsubsection{Correlation analysis with the diagnosis results} \label{sec:correlation} Our results not only identify AD, but also shed light on system design, e.g., what digital biomarkers should be monitored more accurately, which reduces the system's cost and improves the detection accuracy. Specifically, we analyze the correlation between each biomarker and the three diagnostic tasks with ANOVA \cite{st1989analysis}. 
We name those biomarkers with a p-value less than 0.05 as \textit{cretical digital biomarkers}, as they are strongly associated with the diagnostic tasks.
First, the critical digital biomarkers for different diagnostic tasks are different. For example, phone calls, walking, and standing have a strong correlation in differentiating AD/non-AD, yet are not that useful in identifying normal and MCI.  
Moreover, it is extremely hard to differentiate cognitively normal subjects and MCI, as only ``sitting'' is strongly correlated with this diagnosis task. 
This also shows that MCI and cognitively normal subjects exhibit similar activity behaviors.


In addition, we evaluate the impact of gender and age on diagnosis accuracy using the detected digital biomarkers. The results show that the diagnosis accuracy is similar among males (80.55\%) and females (81.82\%).
Moreover, the digital biomarkers are more effective
for the subjects aged 60-70 (88.89\%) and 80-90 (89.47\%) years old, compared with subjects aged 70–80 years old (84.38\%).
We believe these results provide insights into further investigation in AD research.

%% file: 9_discussion.tex
\section{Discussion}
\label{sec:discussion}

{
Here we discuss some future directions of ADMarker. 

\textbf{Improve performance of weakly supervised learning.} Since most subjects/caregivers fail to record activities in detail, the weak labels extracted from the activity
logs for online supervised FL are usually very sparse and limited. In the future, we will develop new approaches that require fewer weak labels during the deployment, e.g., by associating the weak and manually annotated labels during model pre-training on the cloud, or leveraging pseudo-labels automatically generated by applying image classification algorithms to high-quality depth data \cite{xia2013spatio, wang20203dv}. Moreover, we can exploit the temporal dependence of events \cite{jeyakumar2023x} to improve the performance of supervised learning. 

\textbf{Long-term management and interventions with detected biomarkers}. The biomarkers detected by our system could
enable medical experts to investigate how the physical and social interactional etiopathogenesis shapes AD’s manifestation for long-term management and personalized intervention. For example, the duration and intensity of exercise can be prescribed for personalized intervention. Moreover, we could identify critical digital biomarkers that contribute to diagnosis results. For example, phone calls, walking, and standing have a strong correlation in differentiating AD/non-AD (see Section \ref{sec:correlation}). 


\textbf{Ethical and social challenges of real-world deployment.} 
We have obtained informed consent from users (or
their families for AD patients) before the study, where users are fully aware of the types of data to be collected and used. However, there are several areas where we can enhance our clinical deployment process.
First, we will further optimize the hardware design and clinical study protocol to improve the level of user acceptance based on collected user feedback during and after the study. 
Second, although our system preserves users’ data privacy through FL, we could incorporate security measures to further safeguard sensitive information of the system. Finally, although our system achieves reasonably good results in different placements and homes, we may deploy more nodes in other areas of subjects’ homes (e.g., mmWave radar in the bedroom and bathroom) to improve the sensing coverage without compromising users' privacy.

\textbf{Apply to other medical conditions/applications.} Our system can be adapted to the longitudinal monitoring of other chronic diseases that exhibit symptoms related to complex behaviors (e.g., depression and Parkinson’s disease) or other smart home/building applications that require long-term user activity information (e.g., energy management and occupant behavior analysis). In these applications, the weak labels can be obtained through user interfaces embedded in smartphones or homes. For example, marking the time of having lunch automatically labels data during lunch. 
}



%% file: 10_conclusion.tex
\section{Conclusion and Discussion}
\label{sec:conclusion}


We propose ADMarker, the first end-to-end system that integrates multi-modal sensors and new federated learning algorithms for detecting multidimensional AD digital biomarkers in natural living environments. Our system has been deployed in a four-week clinical trial involving 91 elderly participants, which detects various biomarkers in real-world settings and identifies AD with high accuracy.
This study provides key insights into the development of clinically proven digital biomarkers for early AD diagnosis and intervention.